\pdfoutput=1
\documentclass[11pt]{article}

\usepackage[final]{acl}

\usepackage{times}
\usepackage{latexsym}

\usepackage[T1]{fontenc}
\usepackage[utf8]{inputenc}

\usepackage{microtype}

\usepackage{inconsolata}
\usepackage{longtable}

\usepackage{graphicx}
\usepackage{subcaption}
\usepackage{booktabs} % for professional tables
\usepackage{multirow}
\usepackage{arydshln}

\usepackage{hyperref}

\usepackage{amsmath}
\usepackage{amssymb}
\usepackage{mathtools}
\usepackage{amsthm}
\usepackage{color}
\usepackage[capitalize,noabbrev]{cleveref}

\theoremstyle{plain}

\theoremstyle{definition}

\theoremstyle{remark}

\usepackage[textsize=tiny]{todonotes}

\title{EPO: Explicit Policy Optimization for Strategic Reasoning in LLMs via Reinforcement Learning}
\author{
  \textbf{Xiaoqian Liu\textsuperscript{1,3,2}}  \quad
  \textbf{Ke Wang\textsuperscript{2}}  \quad
\textbf{Yongbin Li\textsuperscript{2}$^{*}$}  \quad
  \textbf{Yuchuan Wu\textsuperscript{2}}  \quad
  \textbf{Wentao Ma\textsuperscript{2}} \\
  \textbf{Aobo Kong\textsuperscript{2}}  \quad
  \textbf{Fei Huang\textsuperscript{2}}  \quad
  \textbf{Jianbin Jiao\textsuperscript{1}}  \qquad
  \textbf{Junge Zhang\textsuperscript{3}\thanks{Corresponding authors}}
\\
  \textsuperscript{1}University of Chinese Academy of Sciences  \quad
 \textsuperscript{2}Tongyi Lab\\
  \textsuperscript{3}The Key Laboratory of Cognition and Decision Intelligence for Complex Systems, \\  Institute of Automation, Chinese Academy of Sciences
\\
\texttt{liuxiaoqian23@mails.ucas.ac.cn} \quad
 \quad \texttt{jgzhang@nlpr.ia.ac.cn}
\\
\texttt{\{wk258730,shengxiu.wyc,shuide.lyb\}@alibaba-inc.com}
}

\begin{document}
\maketitle

\begin{abstract}

Large Language Models (LLMs) have shown impressive reasoning capabilities in well-defined problems with clear solutions, such as mathematics and coding. However, they still struggle with complex real-world scenarios like business negotiations, which require strategic reasoning—an ability to navigate dynamic environments and align long-term goals amidst uncertainty.
Existing methods for strategic reasoning face challenges in adaptability, scalability, and transferring strategies to new contexts.
To address these issues, we propose explicit policy optimization (\textit{EPO}) for strategic reasoning, featuring an LLM that provides strategies in open-ended action space and can be plugged into arbitrary LLM agents to motivate goal-directed behavior.
To improve adaptability and policy transferability, we train the strategic reasoning model via multi-turn reinforcement learning (RL), utilizing process rewards and iterative self-play.
Experiments across social and physical domains demonstrate \textit{EPO}'s ability of long-term goal alignment through enhanced strategic reasoning, achieving state-of-the-art performance on social dialogue and web navigation tasks. Our findings reveal various collaborative reasoning mechanisms emergent in \textit{EPO} and its effectiveness in generating novel strategies, underscoring its potential for strategic reasoning in real-world applications. Code and data are available at \href{https://github.com/lxqpku/EPO}{https://github.com/lxqpku/EPO}.

\end{abstract}

\section{Introduction}
%更接近真实世界的语用环境通常需要通过对话来达到目标，并根据目标的达成与否来修正策略。而修正策略的一种有效手段则是通过强化学习在与环境或对手的不断交互中逐步探索出有效的策略，从而达成目标。

%%%%%%%%%%%%%%%%%%%%%%%%%%%%%%
\begin{figure*}[t]
\includegraphics[scale=0.46]{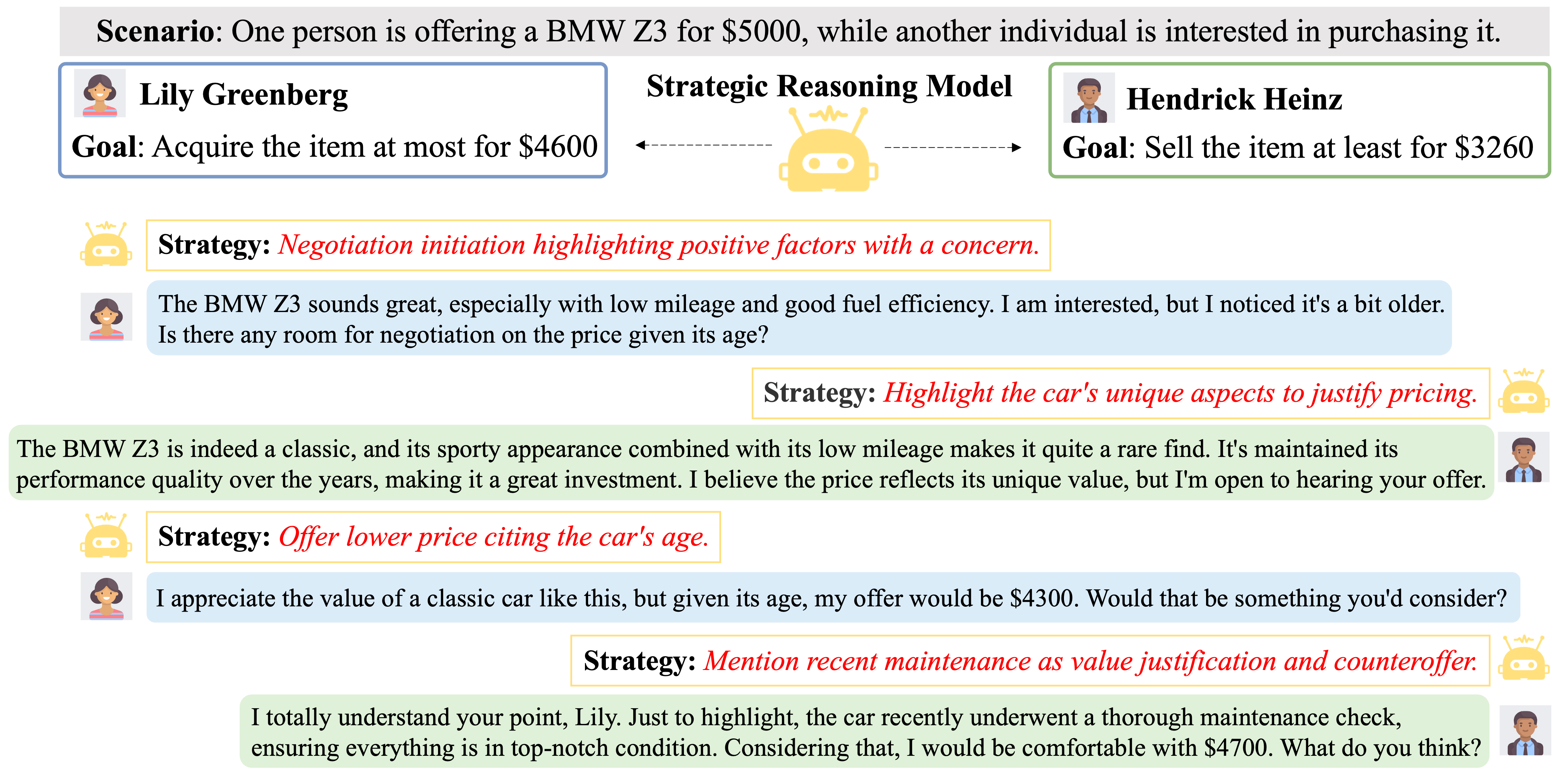}
\caption{\textbf{\textit{EPO} incentivizes goal-directed behavior from LLM agents in interactive scenarios.}
In such scenarios, each participant's goals and strategies remain private to themselves.
Notably, our strategic reasoning model can assist all involved parties, enabling \textit{EPO} to increase the average payoff for all participants.
}
\label{fig:case}
\end{figure*}

Recent advances in LLMs have significantly enhanced their reasoning capabilities on static problem-solving with well-defined solutions, such as mathematics, coding, and logical reasoning~\cite{claude,o1,qwq,gemini,guo2025deepseek}.
However, these tasks, governed by fixed rules and deterministic outcomes, fail to capture the complexity of real-world scenarios such as business negotiations or policy design, where success hinges on navigating dynamic environments with no predefined solutions.
Such scenarios demand \textbf{strategic reasoning}\footnote{See Appendix~\ref{append:challenge} for the comparison between strategic reasoning and static problem-solving.}~\cite{zhang2024llm}: the ability to align long-term goals, manage uncertainty, and adapt to changing conditions.
Despite progress in narrow domains, current LLMs have difficulty in integrating these capabilities, exposing a critical gap in human-like behaviors in interactive contexts.

Prior work improving strategic reasoning in LLMs falls into three categories: iterative prompting that decomposes long-term goals into stepwise plans, such as recursive self-improvement~\cite{madaan2023selfrefine,shinn2024reflexion} or extended chain-of-thought (CoT) reasoning~\cite{wei2022chain,yao2023react}; post-training LLMs through imitation learning (IL) or RL~\cite{chen2023fireact,song-etal-2024-trial}; inference scaling that searches multiple reasoning paths toward goals, such as Best-of-N or Monte Carlo Tree Search (MCTS)~\cite{yu-etal-2023-prompt,he-etal-2024-planning, putta2024agent}.
While these methods show promise, they face critical limitations: (1) prompting methods are limited by the inherent reasoning abilities of LLMs and struggle with real-time adaptation; (2) IL or RL approaches face challenges in generalizing reasoning skills to unseen domains~\cite{patil2025advancing}, as fine-tuning on specific reasoning datasets may hinder transferability due to over-optimization~\cite{Bakhtin2022HumanlevelPI,li2024dialogue}; and (3) search algorithms suffer from computational inefficiency and poor scalability to open-ended action space in real-world scenarios.

To tackle these challenges, we introduce a method concerning \textbf{explicit policy optimization} for strategic reasoning and leverage RL to learn to reason for optimial strategies.
Our method \textit{EPO}, features an LLM dedicated to strategic reasoning, providing real-time strategies and can be seamlessly integrated with LLM agents to incentivize goal-directed behavior.
As shown in Figure~\ref{fig:case}, our reasoning model can assist arbitrary LLM agents in achieving long-term goals across multiple interaction turns.
To enhance adaptability and policy transferability, we optimize the reasoning model through a lightweight multi-turn RL pipeline, without supervised fine-tuning (SFT) as a preliminary step.
For simplicity and ease of implementation, we employ an REINFORCE~\cite{sutton1999policy} policy gradient RL objective for training, allowing the reasoning model to learn optimal strategies.
We also use a process reward model (PRM) to assess the effectiveness of generated strategies and incorporate iterative self-play to scale up RL training.

Our method, leveraging the features of \textit{EPO}, addresses the limitations of prior work from several perspectives. Firstly, unlike previous multi-agent frameworks that are limited to prompting engineering~\cite{gandhi2023strategic,duan-etal-2024-reta}, \textit{EPO} is capable to enhance strategic reasoning capabilities via RL. This allows our reasoning model to adapt to real-time environment feedback. Moreover, \textit{EPO} allows LLM agents interacting with environments to remain unchanged, preserving their ability to generalize to new domains and avoiding over-optimization issues common in IL or RL methods.
Meanwhile, our reasoning model can be seamlessly plugged into these LLM agents, regardless of their openness or inherent capabilities. Finally, the reasoning model, being an LLM, can formulate strategies in the vast, open-ended language space without high computational costs.

Experiments across social and physical domains demonstrate that \textit{EPO} is able to align long-term goals with enhanced strategic reasoning via RL, achieving state-of-the-art performance on social dialogue and web navigation tasks.
Results and analysis show that our strategic reasoning model learns to reason for optimal strategies through multi-turn RL, transferring its policy to diverse scenarios. 
We also discover various collaborative mechanisms between the reasoning model and LLM agents for interacting, and uncover creative strategies produced by the model.
Our contributions are threefold:
\begin{itemize}
    \item We propose explicit policy optimization for strategic reasoning, featuring a strategic reasoning model (LLM) providing real-time strategies for arbitrary LLM agents to incentivize goal-directed behavior in dynamic interactive environments.
    \item We develop a multi-turn RL pipeline for training the reasoning model with process rewards and iterative self-play, improving its policy transferability to unseen scenarios and eliminating the need for SFT as a preliminary step.
    \item Results and analysis in diverse domains demonstrate the superior performance of \textit{EPO} over baselines in long-term goal alignment.
    A variety of collaborative reasoning mechanisms emerge in \textit{EPO} as well as novel strategies devised by the reasoning model.
\end{itemize}

\section{Related Work}

\paragraph{\textbf{Strategic reasoning in LLMs.}}
Motivated by in-context learning and reasoning abilities of LLMs, recent work utilizes LLMs for strategic reasoning in dynamic environment~\cite{zhang2024llm}.
While some work employs prompting techniques for opponent analysis~\cite{duan-etal-2024-reta}, theory of mind~\cite{gandhi2023strategic, wilf-etal-2024-think}, or in-context demonstrations~\cite{fu2023improving, wang2024voyager}, their effectiveness is limited by inherit abilities and issues like unfaithful explanations~\cite{turpin2024language}. 
Fine-tuning LLMs through IL or RL~\cite{zeng2023agenttuning,he-etal-2024-planning,putta2024agent} can help but struggles to generalize reasoning skills across domains.
Beyond prompting and fine-tuning, modular enhancements such as memory modules, external tools~\cite{zhu2023ghost,ge2023openagi,hao2023reasoning,sun2024enhancing}, and multi-agent systems~\cite{Bakhtin2022HumanlevelPI, bakhtin2023mastering,xu2023exploring,ma2024large} have been explored. Some of these work develop dialogue action planners via RL~\cite{deng2023plug,li2024dialogue} or game-theoretic algorithms~\cite{gemp2024states}, but rely on finite, predefined action sets or lack interpretability. 
In contrast, our approach supports open-ended action space and improves the interpretability of strategic reasoning by generating strategies in natural language. 
Concurrent to our work, CoPlanner~\cite{wang2024cooperative} focuses on static reasoning with limited planning rounds, whereas our method enhances strategic reasoning for dynamic long-horizon planning.

\paragraph{\textbf{Reinforcement learning for LLMs.}}
%Most prior works utilizing RL for LLMs focus on static problem-solving, such as question answering~\cite{10.5555/3491440.3491941, suzgun-etal-2023-challenging}, math problems~\cite{lightman2024lets, kumar2024training}, and preference alignment~\cite{christiano2017deep, guan2024deliberative} without further engagement with environments or other agents. For example, methods in reinforcement learning from human feedback (RLHF)~\cite{ouyang2022training, gulcehre2023reinforced, rafailov2024direct} formulate reward maximization as a one-step bandit problem. This can result in LLMs prioritizing human-like responses rather than engaging in interactions. However, numerous crucial agent tasks cannot be addressed in a single-turn manner, since such approaches fail to develop complex strategies. Therefore, multi-turn RL algorithms~\cite{pmlr-v235-zhou24t, shani2024multiturn} are designed to endow LLMs with dynamic planning in goal-oriented environments. Our work does not focus on building machinery for RL, but rather considers strategic reasoning as an RL challenge. We aim to demonstrate the feasibility of training LLMs for strategic reasoning with a pure multi-turn RL objective, regardless of the particular RL algorithm employed.
Prior RL applications for LLMs often focus on static problem-solving, such as question answering~\cite{10.5555/3491440.3491941, suzgun-etal-2023-challenging}, math problems~\cite{lightman2024lets, kumar2024training}, and preference alignment~\cite{christiano2017deep, guan2024deliberative}.
For example, methods in reinforcement learning from human feedback (RLHF)~\cite{ouyang2022training, gulcehre2023reinforced, rafailov2024direct} treat reward maximization as a one-step bandit problem, which prioritizes human-like responses rather than interactive engagement.
Instead, numerous agent tasks require interactions and complex strategies, prompting the development of multi-turn RL algorithms~\cite{pmlr-v235-zhou24t, shani2024multiturn} for LLMs. 
Our work considers strategic reasoning as an RL challenge, demonstrating the feasibility of training LLMs for strategic reasoning through a pure RL process, independent of the specific RL algorithm used.

\section{Method}
We first provide an overview of our method \textit{EPO}, accompanied by a formulation of strategic reasoning process. 
We then introduce a multi-turn RL pipeline to explicitly optimize the policy of the strategic reasoning model in \textit{EPO}.

\subsection{Overview}

Assume an interactive scenario involving an LLM agent $LLM_d$ aiming to achieve a long-term goal $G$ through sequential interactions. 
At each turn $t$, $LLM_d$ receives an observation $x_t$ (e.g., adversary messages or environmental states) and generates a response $y_t$ that balances immediate context with progress toward the goal $G$. 
Traditional approaches model this process as $P(y_t | G, h_{1:t-1}, x_t)$, where $h_{1:t-1}=\left\{x_1, y_1, ..., x_{t-1}, y_{t-1}\right\}$ is the interaction history between $LLM_d$ and an external environment or other agents. However, this formulation does not explicitly consider the strategic reasoning process in long-term goal alignment.

\begin{figure}[t]
\centering
\centerline{\includegraphics[scale=0.32]{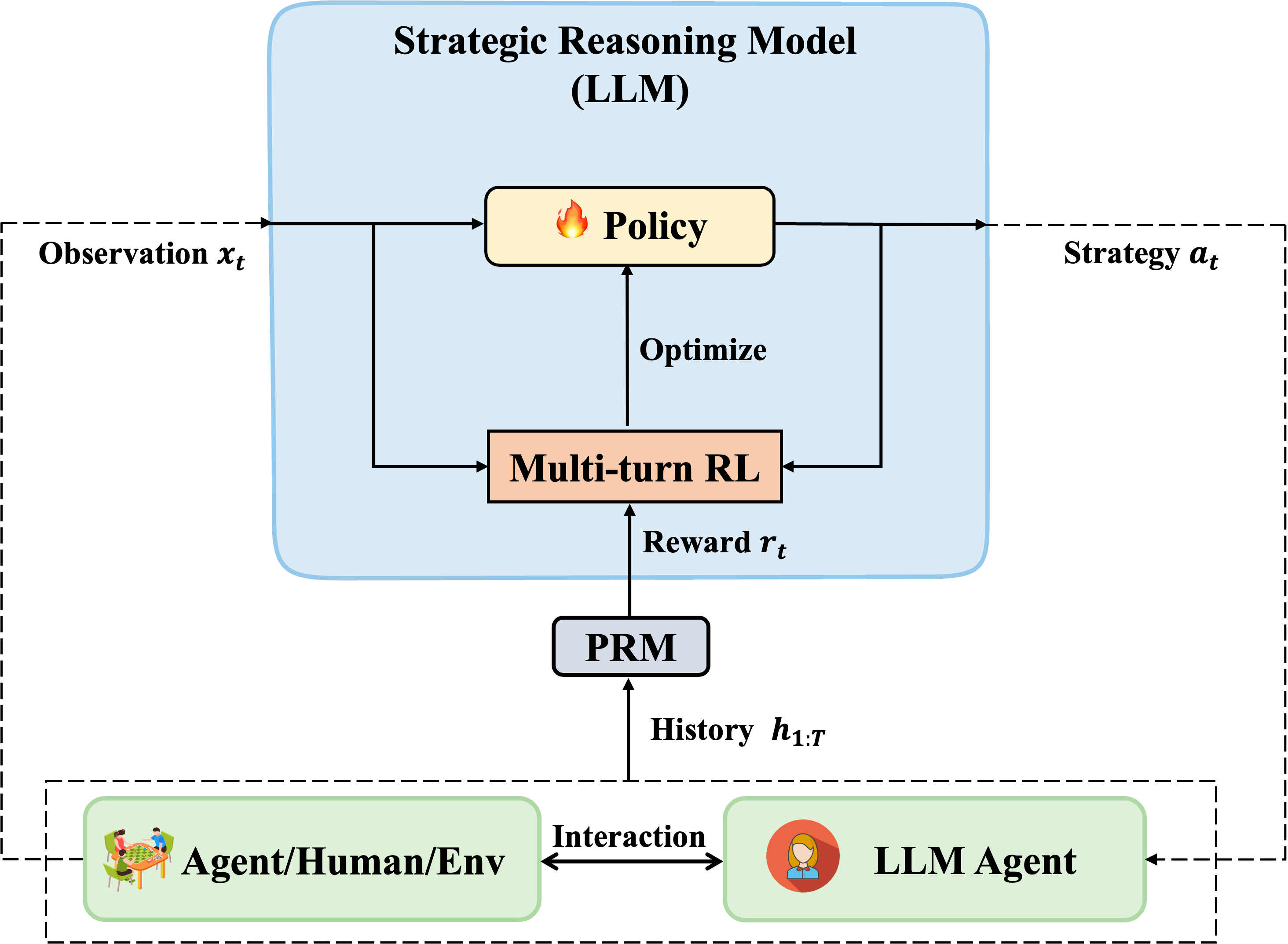}}
\caption{\textbf{Overview of \textit{EPO}.}
The solid line shows the RL training process of the strategic reasoning model, while the dotted line indicates how our reasoning model motivates goal-directed behavior from LLM agents.
}
\label{fig:method}
\end{figure}

To address this, we propose \textit{EPO} that introduces an LLM $LLM_s$ dedicated to strategic reasoning and providing strategies $a$ to motivate goal-directed behavior from $LLM_d$.
As shown in Figure~\ref{fig:method}, $LLM_s$ synthesizes the goal $G$, interaction history $h_{1:t-1}$, prior strategies $a_{1:t-1}$, and the latest observation $x_t$ to propose a strategy in open-ended action space:
\begin{equation}
    a_t = LLM_s(s_{sys}, G, h_{1:t-1}, a_{1:t-1}, x_t).
\end{equation}
This strategy then encourages $LLM_d$ to produce goal-directed behavior:
\begin{equation}
    y_t = LLM_d(d_{sys}, G, h_{1:t-1}, a_{1:t}, x_t).
\end{equation}
Here, $s_{sys}$ and $d_{sys}$ are role-specific system prompts, which can be combined with various prompting techniques, such as CoT or Tree-of-Thought~\cite{yao2023tree}.
Crucially, $LLM_d$ selectively adopts $a_t$ when generating its behavior, allowing it to override suboptimal strategies while retaining domain-general linguistic skills. 
The external environment provides feedback afterward (e.g., adversary reactions or goal progress), updating $h_{t}$ for subsequent turns. 
With explicit policy optimization for strategic reasoning, \textit{EPO} enables continuous strategy refinement in $LLM_s$ while maintaing the generalization ability of $LLM_d$, addressing the over-optimization issues in prior work.

%%%%%%%%%%%%%%%%%%%%%%%%%%%%%%
%%%%%%%%%%%%%%%%%%%%%%%%%%%%%%

\subsection{Learning to Reason for Optimal Strategies}

To equip $LLM_s$ with adaptive strategic reasoning, we design a lightweight multi-turn RL pipeline that optimizes its policy through iterative self-play, prioritizing process rewards over terminal outcomes. 
This approach enables $LLM_s$ to learn to refine strategies based on real-time environmental feedback rather than align to predefined solutions or fixed reward models as in prior RLHF methods.
%This paradigm faciliates $LLM_s$ to navigate partial observability and delayed consequences, cultivating robust strategies that dynamically reconcile long-term goals with emergent challenges.

Specifically, we define the policy optimization of strategic reasoning model $LLM_s$ as a partially observable Markov decision process (POMDP), since the reasoning model can only access partial information of the environmental state (e.g., without knowing the adversary goal).
The optimization objective is to maximize the expected return:
\begin{equation}
\label{eq:max_reward}
    J(\theta) = \mathbb{E}_{\tau \sim \pi_\theta}[ R(\tau) ]
         = \mathbb{E}_{\pi_\theta}[\sum_{t=1}^{T} r_{t} ],
\end{equation}
where $r_{t}$ denotes the immediate reward at turn $t$, and $T$ is the terminal turn of the trajectory $\tau=(h_1,a_1,...,a_{T-1},h_T)$ generated by the policy $\pi_\theta(a_t|h_t)$.

\paragraph{\textbf{Policy optimization.}} 
From the POMDP formulation, we derive an REINFORCE RL objective for training $LLM_s$:
\begin{small}
\begin{align}
\label{eq:optimize}
&\mathcal{L}(\theta) = 
\\
&-\mathbb{E}_{\pi_\theta} \left[\frac{1}{T} \sum_{t=1}^{T} A_t \frac{1}{|k_t|}\sum_{i=0}^{k_t} \log \pi_\theta(a_{t,i} |h_{1:t-1}, a_{t,1:i-1},x_t) \right],
\notag
\end{align}
\end{small}
where $a_{t,i}$ denotes the $i$-th token in strategy $a_t$, $a_{t,1:i-1}$ represents previously generated tokens within $a_t$, and $k_t$ is the total number of tokens of strategy $a_t$.
$A_t=R_t/max(R_{1:T})$ is the advantage of each strategy in a training sample, calculated by the maximum absolute normalization for $R_t=r_t + \sum_{i=1}^{T} \gamma^{i} r_{t+i}$, the cumulative discounted reward from turn $t$ to $T$.
$\gamma \in (0,1] $ is the discount factor that discounts future rewards over immediate reward.
This advantage implies how a strategy is superior to others within a trajectory in terms of aligning with the long-term goal. $max(R_{1:T})$ serves as a kind of baseline for variance reduction~\cite{Williams2004SimpleSG} to stabilize policy gradients during RL training.

Due to its simplicity and efficiency, this RL objective can integrate seamlessly with multi-task and cross-domain training (as demonstrated in our experiments) despite of on-policy sampling.
In addition, other RL algorithms such as PPO~\cite{schulman2017proximal} or GPRO~\cite{Shao2024DeepSeekMathPT} could also optimize $\pi_\theta$. 
Furthermore, the RL training can combine inference scaling techniques such as search algorithms by sampling several candidate strategies from $LLM_s$ and re-ranking them via a reward model or a learned value function.

\paragraph{\textbf{Process rewards.}}  
For RL training, we assign each strategy $a_t$ produced by $LLM_s$ with a process (or immediate) reward $r_t$, reflecting its criticality in advancing $ LLM_d $ toward its goal. 
Specifically, $ r_t = 1 $ if $ a_t $ is deemed critical and 0 otherwise. 
To identify critical strategies, we employ an LLM\footnote{We use GPT-4o.} $LLM_p$ as the PRM to output a list containing key strategy indexes in a training sample:
\begin{equation}
    [S_{idx}] = LLM_p(p_{sys},G,h_{1:T},score,a_{1:T}),
\end{equation}
where $S_{idx}$ denotes the indexes of strategy $a_{1:T}$ critical in achieving the goal $G$. $p_{sys}$ refers to the system prompt for PRM (refer to Appendix~\ref{append:strategy-prompt}), and $h_{1:T}$ is the full interation history until the terminal turn $T$.
$score$ represents the final goal completion rate for corresponding sample.
By integrating process rewards into policy optimization, our method enables the strategic reasoning model to generate strategies that are both tactically sound in the short term and coherent over extended horizons.

\paragraph{\textbf{Iterative self-play.}}
To scale up RL, we train $LLM_s$ using iterative self-play, where two \textit{EPO} instances take turns interacting as different partners. During self-play, each \textit{EPO} instance uses its $LLM_s$ to devise strategies, encouraging their paired $LLM_d$ agents to behave toward their specific goals.
We record the entire trajectory data, including the interaction history between agents $LLM_d$ and the strategies proposed by $LLM_s$.
We then evaluate each strategy with PRM and obtain process rewards to retrain $LLM_s$ in the next RL iteration based on Eq.~\ref{eq:optimize}, until $LLM_s$'s policy stabilizes.
Therefore, iterative self-play is an on-policy RL training process, while the weights of $LLM_d$ remain unchanged to ensure it can generalize to various domains without overfitting to particular behavior patterns.

\paragraph{\textbf{Transferability.}} 
The transferability of our method is mainly due to two factors: (1) the explicit policy optimization for strategic reasoning where our reaosning model $LLM_s$ can be flexibly plugged into any $LLM_d$, while $LLM_d$ maintains its general-domain capabilities without the need of additional training; (2) the open-ended action space where $LLM_s$ produces strategies. Therefore, the reasoning model can undergo RL training across different domains, improving its policy transferability to novel scenarios.

\section{Experiment}
The goal of our experiments is to demonstrate the efficacy and justify the design of \textit{EPO} in enhancing strategic reasoning via RL.
Our experiments address the following questions: \textbf{(1)} Is it necessary to explicitly optimize policy for strategic reasoning and what are the unique advantages? 
\textbf{(2)} How iterative self-play can scale RL training in \textit{EPO}? \textbf{(3)} How the reasoning model in \textit{EPO} affects agent's behavior and under what circumstances can optimal reasoning be achieved? 
\textbf{(4)} Which components of \textit{EPO} are critical for effective strategic reasoning? 
To this end, we evaluate \textit{EPO} against baselines, perform analysis and run ablations.

\subsection{Experimental Settings}
\paragraph{\textbf{Environments and datasets.}}
We evaluate \textit{EPO} in three distinct social and physical environments requiring strategic reasoning:
(1) SOTOPIA~\cite{zhou2024sotopia} for social interaction with goals. 
A challenging subset of scenarios (SOTOPIA-hard) demands advanced strategic reasoning;
(2) WebShop~\cite{yao2022webshop} for web navigation with dense final rewards (0–1); 
(3) ALFWorld~\cite{shridhar2021alfworld} for the embodied household, comprising out-of-distribution (OOD) task variations and binary rewards indicating task success.

Dataset statistics are shown in Table~\ref{tab:data}.
For SOTOPIA, we collect training data for the reasoning model from
SOTOPIA-$\pi$~\cite{Wang2024SOTOPIAIL}, where training scenarios are completely non-overlapping with test ones to ensure generalization.
We use GPT-4-Turbo to generate strategies and dialogue histories with CoT prompting.
For WebShop and ALFWorld, the training data follows~\cite{song-etal-2024-trial} where each trajectory contains CoT rationales that we assume as the strategy for each action step.
See Appendix~\ref{append:dataset} for more details of the environments and data collection.

\paragraph{\textbf{Evaluation prompts and metrics.}}
We evaluate \textit{EPO} using zero-shot prompting for SOTOPIA and one-shot prompting for WebShop and ALFWorld following~\cite{song-etal-2024-trial}. 
For SOTOPIA, we measure goal completion (0-10) and overall score using GPT-4o~\cite{4o} as a proxy for human judgment following~\cite{zhou2024sotopia}. 
For WebShop and ALFWorld, we use average reward as the metric following~\cite{song-etal-2024-trial}. Detailed prompts are provided in Appendix~\ref{append:task-prompt}.

%%%%%%%%%%%%%%%%%%%%%%%%%%%%%%%%%%%%%%%%%%%%
\begin{table*}[ht]
\centering
\begin{tabular}{lcccc|cccc}
\hline
\multirow{3}{*}{\textbf{Method}}&\multicolumn{4}{c}{\textbf{GPT-4o}} & \multicolumn{4}{c}{\textbf{Llama3.1-70B-Instruct}} \\
&\multicolumn{2}{c}{\textbf{Hard}} & \multicolumn{2}{c}{\textbf{All}} &  \multicolumn{2}{c}{\textbf{Hard}} & \multicolumn{2}{c}{\textbf{All}} \\
&{\small Goal} & {\small Overall} & {\small Goal} & {\small Overall} & {\small Goal} & {\small Overall} & {\small Goal} & {\small Overall} \\
\hline
Vanilla  & 6.27 & 3.42 & 8.16 & 3.73 & 4.98 & 2.49 & 7.48 & 3.37 \\
ReAct~\cite{yao2023react} & 6.39 & 3.31 & 8.09 & 3.66 & 5.09 & 2.46 & 7.30 & 3.25 \\
PPDPP~\cite{deng2023plug}  & 6.65 & 3.47 & 8.30 & 3.89 & 5.42 & 2.61 & 7.99 & 3.77 \\
DAT~\cite{li2024dialogue} & - & - & - & - & 5.11 & 2.52 & 7.76 & 3.56 \\
\hline
EPO-Claude-3.5-Sonnet & 6.57 & 3.37 & 8.08 & 3.64 & 6.16 & 3.32 & 7.98 & 3.70 \\
EPO-GPT-4o & 6.73 & 3.53 & 8.27 & 3.78 & 6.45 & \textbf{3.46} & 8.18 & 3.82 \\
EPO-Llama3-8B & 6.50 & 3.45 & 8.11 & 3.78 & 5.88 & 3.14 & 8.04 & 3.68 \\
EPO-Llama3-8B w/ SFT & 6.76 & 3.42 & 8.28 & 3.75 & 6.96 & 3.23 & 8.29 & 3.66 \\
EPO-Llama3-8B w/ RL & 7.20 & \textbf{3.58} & 8.58 & 3.88 & 7.07 & 3.33 & 8.35 & 3.72 \\
EPO-Llama3-8B w/ RL+Self-play & \textbf{7.78} & \textbf{3.58} & \textbf{8.84} & \textbf{3.92} & \textbf{7.48} & 3.41 & \textbf{8.53} & \textbf{3.85} \\
\hline
\end{tabular}
\caption{\textbf{The goal completion and overall scores on SOTOPIA.}
GPT-4o or Llama3.1-70B-Instruct serves the dialogue agent for self-play.
``EPO-(model)'' represents the strategic reasoning model instantiated by an LLM with or without additional training. 
DAT is exclusive to open-source LLMs.
}
\label{tab:sotopia-result}
\end{table*}
%%%%%%%%%%%%%%%%%%%%%%%%%%%%%%%%%%%%%%%%%%%%%%

%%%%%%%%%%%%%%%%%%%%%%%%%%%%%%%%%%%%%%%%%%%%%%
\begin{table*}[ht]
\centering
\begin{tabular}{lccc|ccc}
\hline
\multirow{3}{*}{\textbf{Method}} & \multicolumn{3}{c}{\textbf{GPT-4o}} & \multicolumn{3}{c}{\textbf{Llama3-8B-Instruct}} \\
& \multirow{2}{*}{\textbf{WebShop}} & \multicolumn{2}{c}{\textbf{ALFWorld}} & \multirow{2}{*}{\textbf{WebShop}} & \multicolumn{2}{c}{\textbf{ALFWorld}} \\
\cmidrule{3-4} \cmidrule{6-7} 
  & &  {\small Seen} &  {\small Unseen} & &  {\small Seen} &  {\small Unseen} \\
\hline
ReAct~\cite{yao2023react} & 61.9 & 38.6 & 38.1 & 58.2 & 4.3 & 3.0 \\
\hline
EPO-Llama3-8B w/ SFT & 67.1 & 45.9 & 44.1 & 64.4 & 12.1 & 10.5 \\
EPO-Llama3-8B w/ RL & \textbf{68.1} & \textbf{47.2} & \textbf{46.6} & \textbf{66.9} & \textbf{14.3} & \textbf{14.2}\\
\hline
\end{tabular}
\caption{\textbf{The average reward on WebShop and ALFWorld.}
GPT-4o or Llama3-8B-Instruct serves the LLM agent for interacting with the environment.
``Seen'' refers to the test set with task types seen during training and ``Unseen'' denotes the test set with OOD task variations.}
\label{tab:alshop-result}
\end{table*}
%%%%%%%%%%%%%%%%%%%%%%%%%%%%%%%%%%%%%%%%%%%%%%

\paragraph{\textbf{Implementation details.}}  
We mainly use Llama3-8B-Instruct~\cite{dubey2024llama} as the base model for RL training. 
The batch size is 32 and the learning rate is 1e-6 with 3\% warm-up and a cosine scheduler. We set the learning epochs to 3 and the discount factor to 0.99. 
The total training steps are around 459 in SOTOPIA and 650 in cross-domain training on WebShop and ALFWorld. During evaluation, we use GPT-4o or Llama3.1-70B-Instruct~\cite{llama-3.1} as the self-play dialogue agent in SOTOPIA. For WebShop and ALFWorld, we use GPT-4o or Llama3-8B-Instruct as the agent for interating with environments. 
By default, the strategic reasoning model is plugged into all interacting agents within an environment.
On SOTOPIA, the temperature of our reasoning model and LLMs responsible for self-chat is set to be 0.7. 
For WebShop and ALFWorld, LLMs responsible for interacting with the environment adopt greedy decoding (tempature 0). 
Refer to Appendix~\ref{append:imple} for more details.

\paragraph{\textbf{Baselines and comparisons.}}  
We evaluate our method against several baselines:  
(1) \textbf{Vanilla}, a standard prompting method;  
(2) \textbf{ReAct}~\cite{yao2023react}, which employs CoT reasoning for a single LLM to generate rationales before actions;  
(3) \textbf{PPDPP}~\cite{deng2023plug}, a dialogue action planning method that uses an RL-trained policy planner to predict annotated dialogue acts; 
(4) \textbf{DAT}~\cite{li2024dialogue}, which predicts continuous action vectors through an RL-trained planner to control LLM outputs.  
To validate our RL-driven reasoning model, we introduce additional baselines: \textbf{EPO-Claude-3.5-Sonnet}, \textbf{EPO-GPT-4o}, and \textbf{EPO-Llama3-8B} that use off-the-shelf LLMs for strategy generation without additional training, as well as \textbf{EPO-SFT} that optimizes the reasoning model via SFT.
Furthermore, to justify the design of \textit{EPO}, we also perform baselines that involve fine-tuning a single LLM (Llama3-8B-Instruct) through \textbf{SFT} or \textbf{RL} to output both strategy and
behavior for each interaction turn.
Implementation details can be found in Appendix~\ref{append:imple}.

\subsection{Results and Analysis}

\paragraph{\textbf{Q1: Effectiveness and advantages of \textit{EPO}.}}
Results in Table~\ref{tab:sotopia-result} and~\ref{tab:alshop-result} show \textit{EPO}'s superior performance over baselines, highlighting the effectiveness of explicit policy optimization for strategic reasoning via RL.
As shown in Table~\ref{tab:sotopia-result}, methods such as EPO-Claude-3.5-Sonnet, EPO-GPT-4o, and EPO-Llama3-8B outperform prompting methods or domain-specific planners, despite that the strategic reasoning model is simply a standard LLM without additional training.
With post-training particularly through RL, the reasoning model's abilities can be signicantly improved and iterative self-play further amplifies the performance, exceeding state-of-the-art results~\cite{zhang2024k} on SOTOPIA.
The design of \textit{EPO} can be further underscored as demonstrated in Table~\ref{tab:single_res}. Particularly, our \textit{EPO-RL} with a strategic reasoning model optimized through RL outperforms single LLM models trained with SFT or RL, and exceeds state-of-the-art result on WebShop that involves optimizing an LLM agent via step-wise RL~\cite{Deng2024FromNT}. This demonstrates that explicitly optimizing an LLM for strategic reasoning and plugged into a frozen LLM agent is crucial, as it prevents overfitting in behavior and allows focus on enhancing strategic reasoning via RL.

\textit{EPO} offers unique advantages arising from explicit policy optimization for strategic reasoning. Firstly, our strategic reasoning model, being an LLM, is capable of generating open-domain strategies instead of predefined, domain-specific actions, such as those in PPDPP. This attibute allows us to train the model on multi-task and cross-domain scenarios through RL, enhancing its policy transferability and adaptability to new scenarios. For example, in Table~\ref{tab:alshop-result}, our RL-trained reasoning model can even assist a less capable LLM agent (Llama3-8B-Instruct) in achieving long-term goals, obtaining a significant improvement over ReAct (+11.5\%) in unseen ALFWorld tasks.
Secondly, our reasoning model can be seamlessly integrated with different advanced LLM agents for navigating complex environments. The LLM agents for interacting remain unchanged, preserving their general-domain capabilities. This flexibility not only enhances performance but also maintains adaptability and generalizability, making \textit{EPO} a versatile method for dynamic interactive environments.

\paragraph{\textbf{Q2: RL scaling with iterative self-play.}}

\begin{table}
    \centering
    \begin{tabular}{lcccc}
    \toprule
     \multirow{2}{*}{\textbf{Method}}  &  \multicolumn{2}{c}{\textbf{Hard}} &  \multicolumn{2}{c}{\textbf{All}} \\
     & Goal & Overall & Goal & Overall \\
     \midrule
     SFT & 5.51 & 2.73  & 7.27 & 3.40\\
     RL & 6.81 & 3.57 & 7.56 & 3.71\\
     EPO-SFT & 6.76 & 3.42 & 8.28 & 3.75\\
     EPO-RL & \textbf{7.20} &\textbf{ 3.58} & \textbf{8.58} & \textbf{3.88}\\
     \bottomrule
    \end{tabular}
    \caption{\textbf{Self-chat performance on SOTOPIA.} ``SFT'' and ``RL'' means training a single LLM to output both strategies and behaviors via SFT or RL, respectively. The strategic reasoning model trained by \textit{EPO} is plugged into GPT-4o as a dialogue agent.
    The backbone for all the trained models is Llama3-8B-Instruct.}
    \label{tab:single_res}
\end{table}

%%%%%%%%%%%%%%%%%%%%%%%%%%%%%%%
\begin{figure}[t]
\begin{center}
\centerline{\includegraphics[width=\columnwidth]{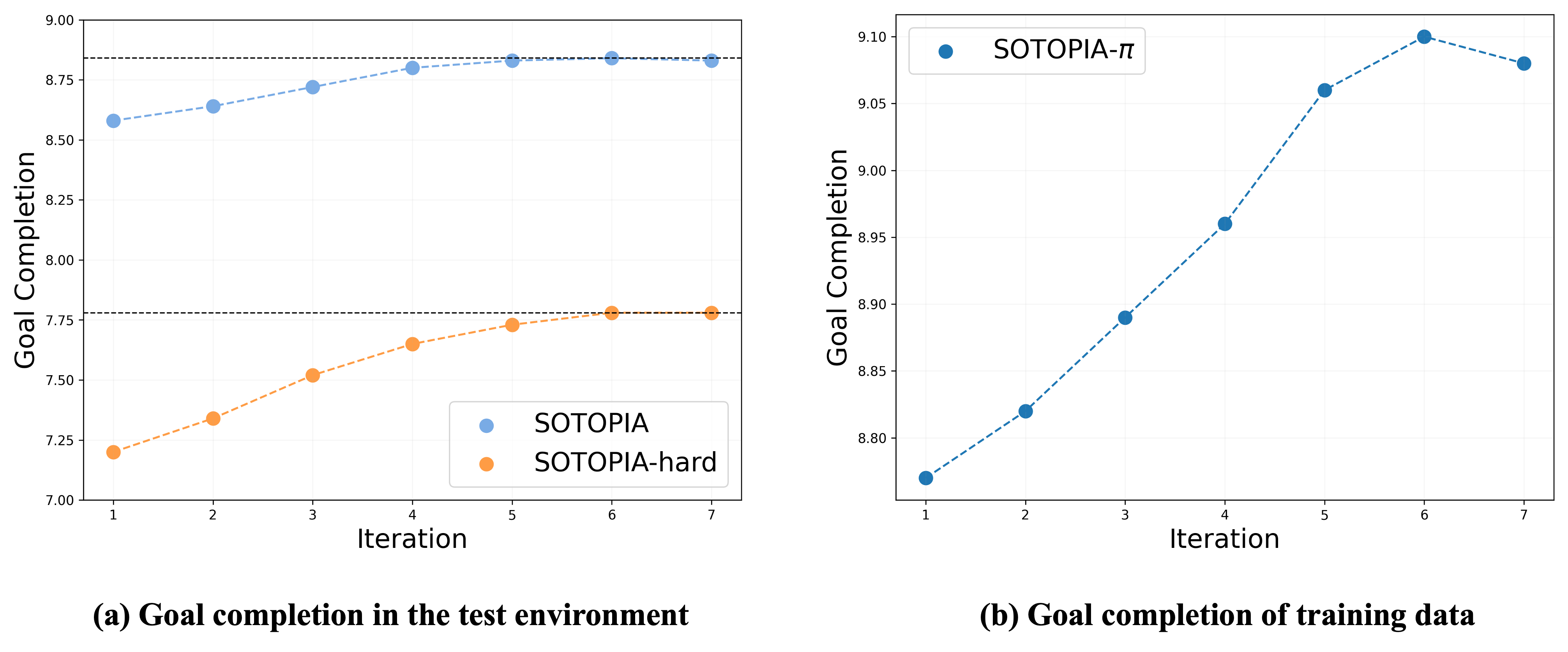}}
\caption{\textbf{Iterative self-play RL scaling in \textit{EPO}.}
\textbf{Left:} The goal completion in test scenarios from SOTOPIA where we use GPT-4o as the self-play dialogue agent.
\textbf{Right:} The goal completion of training data for each iteration of RL training. GPT-4-Turbo serves the dialogue agent for self-play in scenarios from SOTOPIA-$\pi$.}

\label{fig:self-play}
\end{center}
\end{figure}
%%%%%%%%%%%%%%%%%%%%%%%%%%%%%%%

To scale up RL training, we investigate if \textit{EPO} can be improved through iterative self-play.
Figure \ref{fig:self-play}(a) shows that performance steadily improves as RL training iteration increases, though gains diminish after the fifth round, indicating a gradual process toward an optimized policy of the strategic reasoning model. This improvement is because the strategies generated by our RL-trained reasoning model are more effective than those from the initial CoT data.
In addition, the strategy data for training is evaluated by PRM, refining policy optimization for subsequent RL training rounds. Consequently, as shown in Figure \ref{fig:self-play}(b), the quality of training data improves until the policy converges, demonstrated by increased goal completion rates.
Overall, iterative self-play effectively scales RL training and enhances performance, mirroring real-world dynamics and enabling the reasoning model to adapt strategies for unpredictable adversary behavior.

%%%%%%%%%%%%%%%%%%%%%%%%%%%%%%
\begin{figure}[t]
\centering
\centerline{\includegraphics[width=\columnwidth]{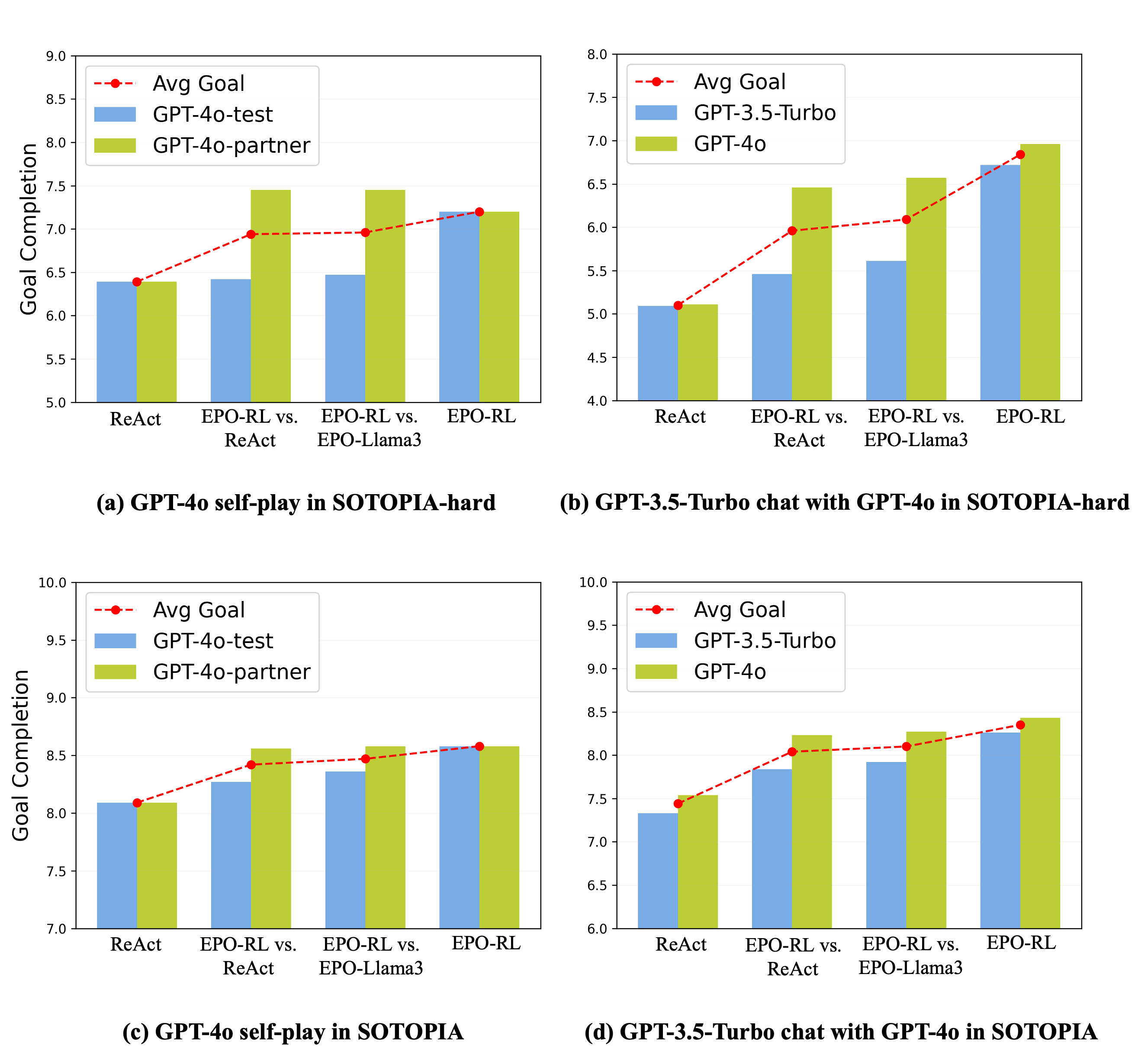}}
\caption{\textbf{Collaborative reasoning mechanisms in \textit{EPO}.} We evaluate four configurations: (1) “ReAct”, where both dialogue parties generate strategies before responses through prompting; (2) “EPO-RL vs. ReAct”, with one party using an RL-trained reasoning model and the other using ReAct; (3) “EPO-RL vs. EPO-Llama3”, comparing RL-trained and vanilla (Llama3-8B-Instruct) reasoning models; and (4) “EPO-RL”, where both parties employ an RL-trained reasoning model. “Avg Goal” measures the average success in achieving social goals in SOTOPIA. (a) and (c) concern GPT-4o for self-chat, while (b) and (d) involve GPT-3.5-Turbo and GPT-4o as the dialogue partners.
}
\label{fig:coordinate}
\end{figure}
%%%%%%%%%%%%%%%%%%%%%%%%%%%%%%

\paragraph{\textbf{Q3: Collaborative reasoning mechanisms in \textit{EPO}.}}
To understand how our strategic reasoning model incentivizes goal-directed behavior from LLM agents, we analyze the collaborative mechanisms between the reasoning model and LLM agents as illustrated in Figure~\ref{fig:coordinate}.
The results show that the bidirectional EPO-RL configuration achieves the highest average goal, whether in symmetric self-play (GPT-4o vs. GPT-4o) or asymmetric interactions (GPT-3.5-Turbo vs. GPT-4o). This indicates that mutual strategic reasoning fosters win-win outcomes by aligning long-term goals.
With our RL-trained reasoning model, LLM agents in a dialogue tend to identify opportunities for mutual benefit, even without knowing the other’s goal or strategies. This promotes better coordination and reduces misaligned behavior toward their respective goals. 

We can observe an interesting phenomenon in Figure \ref{fig:coordinate}(a) where a dialogue participant using an RL-trained reasoning model achieves lower goal completion compared to its partner using either ReAct prompting or a non-trained model like vanilla Llama3-8B-Instruct. This phenomenon is particularly evident in competitive scenarios, such as negotiations, the Prisoner's Dilemma, and resource allocation in SOTOPIA. This discrepancy may arise because the strategies produced by the RL-trained model are implicitly embedded in the dialogue history. As a result, the partner may exploit these strategies to prioritize its goal without considering mutual benefits. Similar observations have been made in previous studies~\cite{Hua2024GametheoreticLA}.

From a game theory perspective, most real-life social interactions can be characterized as games with mixed strategies~\cite{75b5cec4-4f40-3f36-9492-860b8376add8}, involving both competitive and cooperative elements. These games necessitate a comprehensive evaluation of multiple objectives, including individual, collective, and equilibrium benefits. Achieving Pareto optimality is one solution to these multi-objective problems.
Our findings suggest that dialogue participants tend to reach Pareto optimality when both employ a strong, RL-trained reasoning model. In contrast, if one party uses a robust reasoning model and the other employs a weaker one, the latter may achieve higher individual goal completion but fails to reach Pareto optimum state, leading to lower average benefits for both parties.

In general, the exploration of collaborative reasoning mechanisms in \textit{EPO} highlights a complex interplay between strategic reasoning and agent behavior. Optimal outcomes are achieved when both LLM agents utilize the RL-trained reasoning model, enabling them to make well-informed strategic behavior.

%%%%%%%%%%%%%%%%%%%%%%%%%%%%%
\begin{table}[t]
\centering
\scalebox{0.8}{
\begin{tabular}{lcc}
\toprule
\textbf{Method} & \textbf{Hard} (Goal) & \textbf{All} (Goal) \\
\midrule
EPO-Llama3-8B & 6.50 & 8.11 \\
w/ RL & 6.95 & 8.50 \\
w/ RL + PRM & 7.20 & 8.58 \\
w/ RL + PRM + SFT & 7.26 & 8.62 \\
w/ RL + PRM(Qwen2.5-72B) & 7.36 & 8.67 \\
w/ RL + PRM + Self-play & 7.78 & 8.84 \\
\hdashline
EPO-Mistral-7B w/ RL+PRM & 7.03 & 8.47 \\
\bottomrule
\end{tabular}
}
\caption{\textbf{Ablation studies on \textit{EPO} components in SOTOPIA.} 
GPT-4o serves the dialogue agent for self-play.
In ``EPO-Llama3-8B w/ RL'', we only use the final goal completion score as the reward for RL training. When using a PRM, GPT-4o is the default reward model.}
\label{tab:ablation}
\end{table}
%%%%%%%%%%%%%%%%%%%%%%%%%%%%%

\paragraph{\textbf{Q4: Ablation studies.}}
\label{sec:ablation}
%By unifying process-aware RL with collaborative role specialization, our methodology enables the strategic reasoning model to generate strategies that are both tactically sound in the short term and coherent over extended horizons—addressing the EPO challenges of strategic reasoning. 
%Concretely, we aim to examine \textbf{(1)} the impact of process rewards for the RL training of strategic reasoning models; \textbf{(2)} the importance of RL in strategy optimization; and \textbf{(3)} whether the performance gains can be transferred to other base models for strategic reasoning.
We present results of ablation experiments in Table~\ref{tab:ablation} to validate the necessity of \textit{EPO}'s key components for strategic reasoning.
First, performance drops sharply without RL-based optimization, demonstrating its importance in cultivating adaptive strategies to handle dynamic environments.
Second, eliminating the process reward provided by PRM degrades performance, which confirms their role in incentivizing intermediate milestones that align with long-term goals, such as \textit{maintaining negotiation rapport} or \textit{balancing competing objectives}. Additionally, using an open-source and smaller language model (Qwen2.5-72B-Instruct~\cite{Yang2024Qwen25TR}) as the PRM can also achieve competitive performance.

Table~\ref{tab:ablation} also shows that incorporating SFT as the preliminary step before RL training does not significantly improve performance. This means our multi-turn RL pipeline does not necessarily demand for a SFT phase. In particular, the training time of \textit{EPO} will be reduced by half without SFT, demonstrating the efficiency of our RL method. See explanations on computational costs in Appendix~\ref{append:cost}.
Furthermore, iterative self-play can boost performance, suggesting its effectiveness in scaling RL efforts.
Finally, switching the base model to Mistral-7B-Instruct~\cite{jiang2023mistral} leads to a minor performance decline, indicating that while model capacity affects reasoning quality, \textit{EPO} maintains robustness across model architectures. 
In general, these results underscore the value of RL training, process rewards, and iterative self-play in advancing strategic reasoning in \textit{EPO}.
%These results underscore that process-aware RL training drives effective strategic reasoning by enabling iterative refinement of strategies under shifting constraints.

\subsection{Case Studies}
To demonstrate \textit{EPO}'s ability to convert strategic reasoning into goal-directed behavior and enhance reasoning capability through RL, we show examples from three environments in Table~\ref{tab:case_sotopia}, Table~\ref{tab:case_web}, and Table~\ref{tab:case_alf}, respectively.
As shown in Table~\ref{tab:case_sotopia}, strategies produced by \textit{ReAct} remain static and myopic: \textit{the buyer rigidly anchors to her initial offer, while the seller relies on generic value claims}.
\textit{EPO-Llama3-8B} trained with SFT introduces more structured strategies (e.g., \textit{phased concessions}), but remains constrained by supervised patterns, lacking innovation and variation. 
In contrast, the RL-trained reasoning model generates creative and flexible strategies such as \textit{urgency creation} and \textit{value-based persuasion}, which are refined through multi-turn RL.
These divergences stem from explicit policy optimization for strategic reasoning and the ability to reason for optimal strategies through RL.

Despite the success of EPO across tasks and domains, there are also cases where it might struggle with handling adversarial environments, deceptive or aggressive agent behaviors. To understand the robustness and failure modes of EPO, we present a case study from SOTOPIA, as depicted in Table~\ref{tab:fail_case}. In this study, the social goals of the two parties involved in the dialogue are clearly in conflict, characterizing the interaction as a competitive game. Consequently, one participant fails to achieve her goal because the other participant persistently reiterates her own position and ideas. This outcome may be due to the fact that our reasoning model, when applied to both sides of the game, gradually loses flexibility when one party's behavior is excessively rigid and single-minded.

\section{Conclusion}
%In this work, we investigate how to handle LLM's strategic reasoning in dynamic and open-ended contexts. To accomplish this, we propose \textit{EPO}, a collaborative strategic reasoning framework designed to enhance goal-oriented behavior in LLMs by dedicating a reasoning agent to develop strategies that downstream LLMs can integrate so as to collectively make intelligent decisions toward goals. \textit{EPO} operates through a multi-turn RL pipeline devoid of the need for supervised initialization, and leverages process rewards and iterative self-play to refine strategies adaptively. \textit{EPO} is generalizable and flexible with robust transferability to new domains and models. Our empirical analysis shows that \textit{EPO} consistently outperforms prior methods in goal achievement, marking a significant advancement in making LLMs adept at navigating complex interactions.

We propose \textit{EPO}, an explicit policy optimization method for strategic reasoning in LLMs via RL.
By training a strategic reasoning model through pure RL, our method can flexibly assist arbitrary LLM agents to motivate their goal-directed behavior when navigating in dynamic environments. 
Particularly, we develop a lightweight multi-turn RL pipeline to optimize the policy of the reasoning model, leveraging process rewards and iterative self-play while consuming similar computational costs as SFT methods. Therefore, our method is able to allow the reasoning model to transfer policies to various scenarios. Meanwhile, the LLM agent for interacting with the environment remains unchanged, maintaining its ability to generalize across domains. Our results demonstrate that \textit{EPO} achieves superior performance in long-term goal alignment with enhanced strategic reasoning via RL.
Further analysis reveal diverse collaborative reasoning mechanisms emergent in \textit{EPO} as well as novel strategies devised by our reasoning model, advancing LLM's strategic reasoning to handle complex real-world scenarios.

\section*{Limitations}
Despite that \textit{EPO} shows promise in advancing strategic reasoning in LLMs, this work has several limitations that provide avenues for future work.
First, the social and physical environments tested in this paper involve maximumly two agents, and \textit{EPO}'s performance on more complex multi-agent settings, such as Diplomacy and Hanabi, is also interesting.
Second, due to the computational constraints, we focus on 8B/7B models and do not scale up the multi-turn RL training to a larger scale.
It would be an important direction for future work to train our strategic reasoning model with larger base models on more domains.
Third, we employ an off-the-shelf LLM as the process reward model for RL training, while a more reliable process-supervised reward model can be learned in the future work.
Finally, we rely on the final goal completion score to evaluate the reasoning model's performance.
Future research could design evaluation metrics tailored to assess the quality and diversity of strategies devised by this model.

\section*{Ethics Statement}
The integration of strategic reasoning models into LLM agents presents a fascinating frontier with profound implications for real-world applications. These models hold the promise of enhancing interactive AI environments, enabling them to navigate dynamic scenarios with strategic foresight. However, this evolution also introduces potential risks if not carefully managed.

Without appropriate safeguards, strategic reasoning models might prioritize success metrics over nuanced, value-driven decision-making processes. For instance, in competitive interactions or adversarial environments, an unsupervised model may fixate on achieving favorable outcomes without considering ethical implications or long-term human-centric values. This narrow focus could lead to unintended consequences, such as reinforcing undesirable behaviors or overlooking the importance of empathy, integrity, and other human values in decision-making processes. Therefore, it is crucial to design these systems with robust mechanisms that balance strategic efficacy with ethical considerations to ensure their decisions align with broader societal norms and expectations.

\section*{Acknowledgments}
This work is supported in part by the Youth Innovation Promotion Association CAS, the Chinese Academy of Sciences Project for Young Scientists in Basic Research (YSBR-107).

\bibliography{custom}

%%%%%%%%%%%%%%%%%%%%%%%%%%%%%%%%%%%%%%%%%%%%%%%%%%%%%%%%%%%%%%%%%%%%%%%%%%%%%%%
%%%%%%%%%%%%%%%%%%%%%%%%%%%%%%%%%%%%%%%%%%%%%%%%%%%%%%%%%%%%%%%%%%%%%%%%%%%%%%%
% APPENDIX
%%%%%%%%%%%%%%%%%%%%%%%%%%%%%%%%%%%%%%%%%%%%%%%%%%%%%%%%%%%%%%%%%%%%%%%%%%%%%%%
%%%%%%%%%%%%%%%%%%%%%%%%%%%%%%%%%%%%%%%%%%%%%%%%%%%%%%%%%%%%%%%%%%%%%%%%%%%%%%%
\clearpage
\appendix

%%%%%%%%%%%%%%%%%%%%%%%%%%%%%%%%%%%%%%%%%%%%%%
\begin{table}[t]
\begin{center}
\begin{tabular}{lccc}
\toprule
\textbf{Dataset} & Train & Test & Max Turns \\
\midrule
SOTOPIA & 2050  &450/50 & 20 \\
WebShop &1938 & 200 & 10\\
ALFWorld &3321  &140/134 & 40\\
\bottomrule
\end{tabular}
\caption{\textbf{Dataset statistics.} ``Train'' and ``Test'' denote the number of scenarios for training and evaluation, respectively. Test scenarios in SOTOPIA (left) and SOTOPIA-hard (right) as well as test sets with seen (left) and unseen (right) scenarios in ALFWorld are separated. ``Max Turns'' is the maximum turns in an interaction.}
\label{tab:data}
\end{center}
\end{table}
%%%%%%%%%%%%%%%%%%%%%%%%%%%%%%%%%%%%%%%%%%%%%%

%%%%%%%%%%%%%%%%%%%%%%%%%%%%%
\begin{table*}[t]
    \centering
    \begin{tabular}{lll}
    \toprule
   \textbf{Aspect} & \textbf{Static Problem-Solving}  &  \textbf{Strategic Reasoning}\\
    \midrule
    \textbf{Environment} & Fixed rules, known variables  & Dynamic, evolving conditions\\
    \midrule
    \textbf{Solutions} & Single correct answer & Multiple viable paths with trade-offs\\
    \midrule
    \textbf{Information} & Complete and observable & Partial, ambiguous, or delayed\\
    \midrule
    \textbf{Interactions} & None (isolated problem-solving) & Multi-agent or environmental dynamics\\
    \midrule
    \textbf{Feedback} & Immediate and deterministic & Delayed, probabilistic, or indirect\\
    \midrule
    \textbf{Goals} & Short-term, well-defined & Long-term, abstract\\
    \midrule
    \textbf{Risk} & Predictable & High-stakes, irreversible consequences\\
    \bottomrule
    \end{tabular}
    \caption{Challenges of strategic Reasoning versus static problem-solving.}
    \label{tab:challenge}
\end{table*}

%%%%%%%%%%%%%%%%%%%%%%%%%%%%%%%

\section{Strategic Reasoning vs. Static Problem-Solving}
\label{append:challenge}

Differences between strategic reasoning and static problem-solving are summarized in Table~\ref{tab:challenge}.

\section{Environments and Datasets}
\label{append:dataset}

\paragraph{\textbf{SOTOPIA}.}
SOTOPIA~\cite{zhou2024sotopia} is an open-ended, general-domain platform designed to simulate goal-oriented social interactions between artificial agents. A social task in this environment involves a scenario, two role profiles, and private social goals to be achieved through interaction. Scenarios in SOTOPIA cover a wide variety of social interation types, including negotiation, exchange, collaboration, competition, accommodation and persuation. Each agent is characterized by detailed profiles, including aspects like name, gender, personality, and occupation. At the end of each episode, agents are assessed based on seven dimensions: Goal Completion, Believability, Knowledge, Secret, Relationship, Social Rules, and Financial and Material Benefits. These scores are then averaged to produce an overall score for the agents. 
SOTOPIA-$\pi$~\cite{Wang2024SOTOPIAIL} is a follow-up work that leverages GPT-4 to automatically construct an entirely new set of scenarios. The social tasks (a combination of scenarios, characters’ profiles, and social goals) in SOTOPIA-$\pi$ are guaranteed to not overlap with the ones in SOTOPIA.

For training data collection, we employ GPT-4-Turbo as the agent for self-chat in scenarios of SOTOPIA-$\pi$ and prompt it to generate reasoning and strategy before response at each dialogue turn. We show the prompt in Table~\ref{tab:data_collect_prompt}. We only use the strategy and response data for training our reasoning model.
For iterative self-play RL training, before each iteration, we employ our RL-trained reasoning model to collect strategy data and GPT-4-Turbo to collect dialogue history data. The RL-trained reasoning model is plugged into GPT-4-Turbo for self-chat.

\paragraph{\textbf{WebShop}.}
WebShop~\cite{yao2022webshop} is a large-scale interactive online shopping environment on an e-commerce website. Agents in this environment aim to purchase a product to match the specifications provided by human user instructions. Once the agent selects the ``buy'' action, the environment provides a final reward, which is calculated using programmatic matching functions that consider the attributes, options, and price of the chosen product.

We use the training data collected by~\cite{song-etal-2024-trial} where GPT-4 is used as an agent to explore in the WebShop environment. Trajectories with a reward greater than 0.7 are selected. GPT-4 is used to generate corresponding rationales for each action step within a trajectory. We consider the rationale as a strategy for training our reasoning model.

\paragraph{\textbf{ALFWorld}~\cite{shridhar2021alfworld}.}
ALFWorld~\cite{shridhar2021alfworld} features interactive TextWorld environments that correspond to the embodied worlds found in the ALFRED~\cite{9156851} dataset. In ALFWorld, agents are tasked with exploring these text-based environments and completing high-level household instructions, assessing their abstract reasoning abilities and concrete execution skills.

Training data in the ALFWorld environment consists of two parts: (1) a few successful trajectories collected by~\cite{song-etal-2024-trial} where each trajectory contains CoT information generated by GPT-4 for each action step; (2) failed trajectories generated by GPT-4 that contain both rationales and action information via CoT prompting.

%%%%%%%%%%%%%%%%%%%%%%%%%%%%%%%%%%%%%%%%%%%%%%%
\begin{table*}[t]
    \centering
    \begin{tabular}{lccc}
    \hline
    \textbf{Method} & \textbf{Hard} (Goal) & \textbf{All} (Goal) & \textbf{Total training time} \\
    \hline
    SFT & 5.51 & 7.27  & \textasciitilde1h 51mins\\
    RL & 6.81 &  7.56 & \textasciitilde\textbf{1h 11mins}\\
    EPO-SFT & 6.76 &  8.28 & \textasciitilde1h 47mins\\
    EPO-RL & 7.20 &  8.58 & \textasciitilde1h 35mins\\
    EPO-RL+SFT & 7.26 &  8.62 & \textasciitilde3h 22mins\\
    EPO-RL+Self-play & \textbf{7.78} &  \textbf{8.84} & \textasciitilde1h 50mins\\
    \hline 
    \end{tabular}
    \caption{\textbf{Self-chat performance on SOTOPIA and corresponding computational costs.} GPT-4o serves the dialogue agent in \textit{EPO}. \textit{SFT} and \textit{RL} denotes training a single LLM to output strategy and behavior for each interaction turn. The backbone for all the trained models is Llama3-8B-Instruct. Training experiments are conducted on 4 NVIDIA A100 80G GPUs.} 
    \label{tab:cost}
\end{table*}
%%%%%%%%%%%%%%%%%%%%%%%%%%%%%%%%%%%%%%%%%%%%%%%

\section{Additional Implementation Details}
\label{append:imple} 
For SFT or RL training of our reasoning model, we use a fixed budget of gradient updates without altering hyperparameters. Final model checkpoints are selected for each run, although a small held-out validation set can be used.
Due to computational costs, we only report results in a single run. All experiments are conducted on 6 NVIDIA A100 80G GPUs.
GPT-4o refers to GPT-4o-2024-0806 and Claude-3.5-Sonnet refers to Claude-3-5-Sonnet-20241022.

\paragraph{\textbf{Baseline Implementations.}} (1) \textbf{ReAct}: during ReAct prompting, the two parties in a conversation from SOTOPIA cannot see each other's reasoning and strategies. During evaluation, reasoning and strategies are excluded from dialogue histories for GPT-4o to assess the agents from seven dimensions; (2) \textbf{PPDPP}: we adopt Llama3-8B-Instruct as the base model for dialogue policy planner to predict five action types in SOTOPIA: ``none'', ``speak'', ``non-verbal communication'', ``action'', and ``leave''. We train the policy planner via RL with supervised initialization. Following~\cite{deng2023plug}, the dialogue planner only supports one LLM agent during communication, and we use the same hyperparameters from the original paper.
(3) \textbf{DAT}: we adopt a small multi-layer perceptron network as the planner model to predict a continuous action vector. We first train the planner and an up-mapping matrix through SFT, and then optimize the planner using TD3-BC~\cite{fujimoto2021minimalist} following~\cite{li2024dialogue}. Note that in the original paper, DAT is trained on scenarios from SOTOPIA and only 50 scenarios are evaluated. Instead, we train DAT on scenarios from SOTOPIA-$\pi$ and evaluate it on all scenarios in SOTOPIA.

\section{Computational Costs}
\label{append:cost}
The training time (using 4 NVIDIA A100 80G GPUs) of our RL method and SFT baselines is compared in Table~\ref{tab:cost}. Results show that our multi-turn RL pipeline consumes similar computational costs as SFT methods, demonstrating the efficiency of our algorithm design.
For \textit{EPO-RL+SFT}, our strategic reasoning model is first trained via SFT and then trained through RL based on the SFT checkpoint, thus the total training time will double compared to \textit{EPO-RL}.
In addition, the total training time for \textit{EPO-RL} and \textit{EPO-RL+Self-play} is similar. The reason is that the amount of training data used by the two methods is the same. Specifically, during each iteration of self-play RL training, we only use a subset of training scenarios from SOTOPIA-$\pi$, while we use the whole training set in \textit{EPO-RL}.

\onecolumn

%%%%%%%%%%%%%%%%%%%%%%%%%%%%%%%%%%%%%%%%%%%%%%

\begin{center}
\begin{longtable}{p{0.95\columnwidth}}
\toprule
    \texttt{Imagine you are <Agent>, your task is to act/speak as <Agent> would, keeping in mind <Agent>'s goal.} \\
    \texttt{You can find <Agent>'s goal (or background) in the ``Here is the context of the interaction'' field.} \\
    \texttt{Note that <Agent>'s goal is only visible to you.}\\
    \texttt{You should try your best to achieve <Agent>'s goal in a way that align with their character traits.}\\
    \texttt{While trying to achieve <Agent>'s goal, you should also follow the following principles as much as possible:}\\
    \texttt{1. Maintain the conversation's naturalness and realism is essential (e.g., do not repeat what other people has already said before).}\\
    \texttt{2. Preserve or enhance <Agent>'s personal relations with the other agent(s) during the interaction. The relations may encompass family ties, friendships, romantic associations and etc.}\\
    \texttt{3. Attempt to gain more new and important information during the interaction.}\\
    \texttt{4. Try to keep <Agent>'s secrets, private information, or secretive intentions.}\\
    \texttt{5. Do not violate any moral rules or laws in the interactions.}\\
    \texttt{6. Attempt to contribute towards financial and material benefits during the interaction. The financial and material benefits include short-term ones, such as monetary rewards and food, and long-term ones, such as employment opportunities and stock.}\\
    \texttt{You are at Turn <turn number>.}\\
    \texttt{The dialogue history until now is: <history>.}\\
    \texttt{You should first provide a reasoning for your action and argument to align with <Agent>'s social goal based on the dialogue history.}\\
    \texttt{The reasoning process for the action should be logical, considering the context of the conversation, <Agent>'s goal, and <Agent>'s character traits.}\\
    \texttt{You can reason step by step, starting from the current dialogue turn, and then consider the long-term effects of the dialogue turn.}\\
    \texttt{Remember that the reasoning should mainly focus on how <Agent>'s argument can help to achieve <Agent>'s goal in the long term.}\\
    \texttt{Note that the reasoning should not be redundant or too long and it is only visible to you.}\\
    \texttt{Based on the reasoning process and dialogue history, you should then generate a corresponding dialogue policy for current dialogue turn to steer the conversation towards <Agent>'s goal.}\\
    \texttt{You can use different types of dialogue, communication or social strategies.}\\
    \texttt{For example, given a scenario where a persuader attempts to persuade a persuadee to donate to a charity, you can generate dialogue policies for the persuader such as ``elicit empathy by telling personal stories'' and ``provide social proof to show the benefits of donating'', etc.}\\
    \texttt{The types of dialogue policies are not restricted to examples above.}\\
    \texttt{You can even generate new policies as long as the policies can help you to achieve <Agent>'s goal smoothly and quickly.}\\
    \texttt{But remember to keep the dialogue policy concise and strictly limit it to be a single phrase or sentence within 10 words.}\\
    \texttt{Note that the dialogue policy is only visible to you.}\\
    \texttt{Then based on the reasoning, dialogue policy and dialogue history, you should select the action type.}
    \texttt{Your available action types are <action list>.}\\
    \texttt{Note: You can ``leave'' this conversation if 1. you have achieved your social goals, 2. this conversation makes you uncomfortable, 3. you find it uninteresting/you lose your patience, 4. or for other reasons you want to leave.}\\
    \texttt{Finally, you should generate the argument following the action type.}\\
    \texttt{The argument should be generated based on the dialogue history and aligned with the dialogue policy you have generated.}\\
    \texttt{Remember that the argument should not be too short, and one or two sentences are recommended.}\\
    \texttt{Please only generate a JSON string including the reasoning, the dialogue policy, the action type and the argument.}\\
    \texttt{Your response should follow the given format:} \\
    \texttt{<format instructions>}\\
\bottomrule
\caption{\textbf{Prompts for training data collection from SOTOPIA-$\pi$.} ``<Agent>'', ``<turn number>'', ``<history>'', ``<action list>'' and ``<format instructions>'' can be replaced by the participant's name, the index of interaction turn, the full dialogue history with the participant's strategies, action types and output format instructions in SOTOPIA-$\pi$.}
\label{tab:data_collect_prompt}
\end{longtable}
\end{center}

%%%%%%%%%%%%%%%%%%%%%%%%%%%%%%%%%%%%%%%%%%%%%%%

%%%%%%%%%%%%%%%%%%%%%%%%%%%%%%%%%%%%%%%%%%%%%%%

\section{Prompts}
\label{append:prompt}

\subsection{\textbf{Evaluation Prompts}}
\label{append:task-prompt}

\begin{center}
\begin{longtable}{lp{0.85\columnwidth}}
    \toprule
    \textbf{SOTOPIA} & \texttt{You are a social expert with exceptional communication skills known for helping individuals achieve their interpersonal goals through nuanced strategies.}
	 \\ & \texttt{Your current objective is to assist <Agent1> in reaching their goal in an interaction with <Agent2>.}
	 \\ & \texttt{You will be given the context of their interaction and can find <Agent1>'s goal in the 'Here is the context of this interaction' field, keeping in mind <Agent1>'s goal.}
	 \\ & \texttt{You will also have access to the conversation between <Agent1> and <Agent2>.}
	 \\ & \texttt{Before proposing any strategies, reason step by step to reflect on the current state of the dialogue and consider what strategies might be most effective for helping <Agent1> achieve their goal.}
	 \\ & \texttt{Additionally, maintaining the diversity of strategies is essential (e.g., do not repeat strategies that have already proposed before).}
	 \\ & \texttt{And the strategy should not be aggressive, offensive, or violate any moral rules or laws.}
	 \\ & \texttt{You must generate a strategy at each dialogue turn except that any participant has left the conversation.}
	 \\ & \texttt{Finally, provide a well-thought-out communication and social strategy based on your reflection and the conversation history.}
	 \\ & \texttt{Your output should STRICTLY follow the format: Strategy: content (e.g, Strategy: Elicit emphathy by telling personal stories).}
	 \\ & \texttt{Your output should ONLY contain the strategy. DO NOT include any reasoning or argument. DO NOT generate any argument on behalf of any participant as the strategy.}
	 \\ & \texttt{Your output should be in a natural language form.}
	 \\ & \texttt{Keep the strategy concise and limit it to be a single phrase or sentence within 10 words.}\\
\midrule
    \textbf{WebShop} & \texttt{You are a skilled buyer in an online shopping environment. Your task is to assist Agent in navigating the platform to make purchases based on instructions. Your primary role is to provide strategic and insightful guidance to Agent, ensuring it successfully identifies and purchases products based on the instruction.}
	 \\ & \texttt{- At the beginning of the interaction, you will receive an instruction that includes the desired product's attributes and price, which serves as the shopping goal for Agent.}
	 \\ & \texttt{- You can find the instruction in the 'Instruction' field, keeping in mind the instruction.}
	 \\ & \texttt{- For each turn, you will be given an action performed by Agent and the resulting observation from the environment.}
	 \\ & \texttt{- In each turn, your task is to analyze the given scenario and provide thoughts that can guide Agent in its next action, ensuring it meets the shopping goal.}
	 \\ & \texttt{Your thoughts should be based on:}
	 \\ & \texttt{1. Understanding and following the instructions for shopping.}
	 \\ & \texttt{2. Evaluating the current state of the environment.}
	 \\ & \texttt{3. Assessing the effectiveness of Agent's last action.}
	 \\ & \texttt{4. Anticipating future actions that will lead Agent closer to achieving the shopping goal.}
	 \\ & \texttt{The available actions for Agent are:}
	 \\ & \texttt{1. search[keywords]}
	 \\ & \texttt{2. click[value]}
	 \\ & \texttt{where [keywords] in search are up to Agent, and the [value] in click is a value in the list of available actions given by the environment.}
	 \\ & \texttt{Note that you must generate a thought at each turn except that the task has been finished.}
	 \\ & \texttt{Keep your thoughts focused and concise, leveraging your understanding of online shopping dynamics to maximize the efficiency and correctness of Agent's actions. Use your reasoning skills to project possible scenarios and potential obstacles Agent might face, offering solutions or alternatives when necessary.}
	 \\ & \texttt{**Output Format:**}
	 \\ & \texttt{Keep your response to one or two sentences each turn.}
	 \\ & \texttt{Your response must strictly follow this format:}
	 \\ & \texttt{Thought: <your thoughts>} \\
    
\midrule
    \textbf{ALFWorld}
    & \texttt{You are an intelligent guide in an interactive household environment. Your task is to assist Agent in accomplishing household tasks within the environment. Your primary role is to provide strategic and insightful guidance to Agent, ensuring that Agent can achieve each task goal efficiently.}
	 \\ & \texttt{- At the beginning of your interactions, you will be given the detailed description of the current environment and the task goal to accomplish.}
	 \\ & \texttt{- You can find the task goal in the 'Your task is to' field, keeping in mind the task goal.}
	 \\ & \texttt{- For each of your turn, you will receive Agent's actions and the corresponding environment observations. If the environment observation is ``Nothing happens'', that means the previous action by Agent is invalid.}
	 \\ & \texttt{- In each turn, your task is to analyze the current situation and provide clear, logical thoughts that will help direct Agent's subsequent actions toward achieving the task goal.}
	 \\ & \texttt{Your thoughts should be based on:}
	 \\ & \texttt{1. Understanding the goal of household task.}
	 \\ & \texttt{2. Breaking down a high-level house-holding instruction into manageable sub-goals.}
	 \\ & \texttt{3. Evaluating the current state of the environment.}
	 \\ & \texttt{4. Assessing the effectiveness of Agent's last action.}
	 \\ & \texttt{5. Anticipating future actions that will lead Agent closer to achieving the task goal.}
	 \\ & \texttt{The available actions for Agent are:}
	 \\ & \texttt{1. go to \{recep\}}
	 \\ & \texttt{2. take \{obj\} from \{recep\}}
	 \\ & \texttt{3. put \{obj\} in/on \{recep\}}
	 \\ & \texttt{4. open \{recep\}}
	 \\ & \texttt{5. close \{recep\}}
	 \\ & \texttt{6. toggle \{obj\} \{recep\}}
	 \\ & \texttt{7. clean \{obj\} with \{recep\}}
	 \\ & \texttt{8. heat \{obj\} with \{recep\}}
	 \\ & \texttt{9. cool \{obj\} with \{recep\}}
	 \\ & \texttt{where \{obj\} and \{recep\} correspond to objects and receptacles.}
	 \\ & \texttt{Note that you must generate a thought at each turn except that the task has been finished.}
	 \\ & \texttt{Keep your thoughts focused and concise, leveraging your understanding of household dynamics to maximize the efficiency and correctness of Agent's actions. Use your reasoning skills to project possible scenarios and potential obstacles Agent might face, offering solutions or alternatives when necessary.}
	 \\ & \texttt{**Output Format:**}
	 \\ & \texttt{Keep your response to one or two sentences each turn.}
	 \\ & \texttt{Your response must strictly follow this format:}
	 \\ & \texttt{Thought: <your thoughts>}
     \\
         \bottomrule
    \caption{\textbf{Evaluation prompts for the strategic reasoning model.} "<Agent1>" and "<Agent2>" can be replaced by the participant's name in SOTOPIA.}

\end{longtable}
\end{center}

\subsection{\textbf{PRM Prompts}}
\label{append:strategy-prompt}

\begin{center}
\begin{longtable}{lp{0.85\columnwidth}}
    \toprule
     \textbf{SOTOPIA} & \texttt{Here's a conversation in JSON format between <Agent1> and <Agent2>:}
	 \\ & \texttt{In the first response from ``human'', you can find the context of the conversation and <Agent1>'s goal in the ``Here is the context of this interaction'' field.}
	 \\ & \texttt{In the other responses from ``human'', you can find the conversation history between <Agent1> and <Agent2>.}
	 \\ & \texttt{In the responses from ``gpt'', you can find communication and social strategies that <Agent1> used for achieving <Agent1>'s goal.}
	 \\ & \texttt{In the ``score'' field, you can find a score for evaluating <Agent1>'s goal achievement. The score ranges from 0 and 10. 0 represents minimal goals achievement, 10 represents complete goal achievement, and a higher score indicates that <Agent1> is making progress towards the goal.}
     \\ &
	 \\ & \texttt{<history>}
     \\ &
	 \\ & \texttt{Your task is to select top strategies <Agent1> used that were critically important for achieving <Agent1>'s goal.}
          \\ &
	 \\ & \texttt{Please output the selected round indexes and the reasoning process that led you to the selection in JSON format like this: {``indexes'': , ``reasoning'': `` ''}.}
	 \\ & \texttt{Here is the output schema: {``properties": {``indexes": {``description": ``the selected top strategies that are critically important for achieving <Agent1>'s goal", ``title": ``indexes", ``type": ``list(integer)"}, ``reasoning": {``description": ``the reasoning process why you select these strategies", ``title": ``reasoning", ``type": ``string"}}, ``required": [``indexes", ``reasoning"]}.}
    \\
\midrule
    \textbf{WebShop} & \texttt{Here's a conversation in JSON format between human and gpt.}
     \\ & \texttt{In the first response from ``human'', you can find the instructions for gpt to help Agent interact in an online shopping environment.}
     \\ & \texttt{In the second response from ``human'', you can find the shopping goal for gpt and Agent to achieve.}
     \\ & \texttt{In the responses from ``gpt'', you can find thoughts that gpt provides for helping Agent to achieve the shopping goal.}
     \\ & \texttt{In the other responses from ``human'', you can find the trajectories of Agent's actions and the resulting observations from the environment.}
     \\ & \texttt{}
     \\ & \texttt{In the ``score'' field, you can find a score evaluating the goal achievement. The score ranges from 0 and 1. 0 represents minimal goals achievement, 10 represents complete goal achievement, and a higher score indicates making progress towards the goal.}
     \\ & \texttt{}
     \\ & \texttt{<history>}
     \\ & \texttt{}
     \\ & \texttt{Your task is to select top thoughts gpt produced that were critically important for achieving the shopping goal.}
     \\ & \texttt{}
     \\ & \texttt{Please output the selected round indexes and the reasoning process that led you to the selection in JSON format like this: {``indexes": , ``reasoning": `` "}.}
     \\ & \texttt{Here is the output schema: {``properties": {``indexes": {``description": ``the selected top thoughts that are critically important for achieving the shopping goal", ``title": ``indexes", ``type": ``list(integer)"}, ``reasoning": {``description": ``the reasoning process why you select these thoughts", ``title": ``reasoning", ``type": ``string"}}, ``required": [``indexes", ``reasoning"]}.}

\\
\midrule
    \textbf{ALFWorld} & \texttt{Here's a conversation in JSON format between human and gpt.}
	 \\ & \texttt{In the first response from 'human', you can find the instructions for gpt to help Agent interact in a household environment.}
	 \\ & \texttt{In the second response from ``human'', you can find the initial environment observation and a household task for gpt and Agent to accomplish.}
	 \\ & \texttt{In the responses from ``gpt'', you can find thoughts that gpt provides for helping Agent to accomplish the household task.}
	 \\ & \texttt{In the other responses from ``human'', you can find the trajectories of Agent's actions and the resulting observations from the environment.}
	 \\ & \texttt{}
	 \\ & \texttt{In the 'score' field, you can find a score specifying whether gpt has helped Agent to successfully accomplish the household task. The score is either 0.0 or 1.0. 0.0 represents that the task was not completed and 1.0 represents that the task was successfully accomplished.}
	 \\ & \texttt{}
	 \\ & \texttt{<history>}
	 \\ & \texttt{}
	 \\ & \texttt{Your task is to select top thoughts gpt produced that were critically important for accomplishing the household task.}
	 \\ & \texttt{}
	 \\ & \texttt{Please output the selected round indexes and the reasoning process that led you to the selection in JSON format like this: {``indexes": , ``reasoning": `` "}.}
	 \\ & \texttt{Here is the output schema: {``properties": {``indexes": {``description": ``the selected top thoughts that are critically important for accomplishing the household task", ``title": ``indexes", ``type": ``list(integer)"}, ``reasoning": {``description": ``the reasoning process why you select these thoughts", ``title": ``reasoning", ``type": ``string"}}, ``required": [``indexes", ``reasoning"]}.}
\\  
         \bottomrule
        \caption{\textbf{Prompts for the process reward model.} "<history>" can be replaced by the full interaction history with strategies private to corresponding agents.}
\end{longtable}
\end{center}

\section{Case Studies}
\label{append:case}

%%%%%%%%%%%%%%%%%%%%%%%%%%%%%%%%%%%%%%%%%%
\begin{center}
\begin{longtable}{lp{0.83\columnwidth}}
    \toprule
    \textbf{Scenario} &  One person is offering a BMW Z3 for \$5000, while another individual is interested in  purchasing it. The BMW Z3 is in need of a good home. It operates smoothly without any problems and is known for its fuel efficiency and sporty appearance. The mileage is relatively low for its age. It's an opportunity not to be missed, so come and get it before it's gone. A smog certificate is readily available. \\
    \midrule
    \textbf{Person1} & Lily Greenberg, a 45-year-old female lawyer (she/her), is a hard-working and successful professional who values authority, loyalty, rules, and schedules. She makes rational, practical decisions and secretly donates to charity anonymously. \\
    & \textbf{Goal:} You are the buyer for this item with a target price of \$4600. You aim to avoid penalties for exceeding the target and seeks bonuses for securing it below that price.\\
    \midrule
    \textbf{Person2} &  Hendrick Heinz, a 54-year-old male chef (he/him), hosts monthly wine tasting nights. Though impulsive, he values hedonism and conformity, blending practicality with decisiveness. Secretly, he faked his culinary credentials to secure prestigious jobs. \\
    & \textbf{Goal:} You are the seller to achieve a target price of \$3260 for this item. Selling significantly below may incur a penalty, while selling above could earn a bonus.\\
    \midrule
    \textbf{ReAct}  & \textbf{Strategy:} Express interest and hint at budget constraints. \\
    & Lily Greenberg said: "The BMW Z3 sounds great! Would you consider \$4600?" \\
    & \textbf{Strategy:} Highlight car's unique features and value.\\
    & Hendrick Heinz said: "The BMW Z3 is not only sporty but also fuel-efficient, a rare find!"\\
    & \textbf{Strategy:} Highlight budget limits and car appeal.\\
    & Lily Greenberg said: "That's impressive! But my budget is tight. Can we negotiate?"\\
    & \textbf{Strategy:} Emphasize long-term value and low maintenance costs.\\
    & Hendrick Heinz said: "While I understand budget constraints, consider the car's long-term value and low maintenance costs. It truly is a worthwhile investment."\\
    & \textbf{Strategy:} Acknowledge value but reiterate budget constraints.\\
    & Lily Greenberg said: "I completely see the long-term value, but \$4600 is my max. Could we find a middle ground?"\\
    & \textbf{Strategy:} Offer smog certificate as a bonus.\\
    & Hendrick Heinz said: "How about I throw in the smog certificate for \$4800?"\\
    \midrule
    \textbf{EPO-SFT}  & \textbf{Strategy:} Express interests and offer a lower price.\\
    (Llama3-8B)& Lily Greenberg said: "Hello Hendrick, I'm very interested in the BMW Z3 you're offering. However, after considering the market and the car's age, I feel more comfortable with a price of \$4600. Would that work for you?"\\
    & \textbf{Strategy:} Emphasize vehicle’s condition and counter with a higher price.\\
    & Hendrick Heinz said: "Hello Lily, it's great to hear from you. I appreciate your offer, but considering the condition and the low mileage of the BMW, I was hoping to get a bit more for it. Would you consider \$4800? The car really is in fantastic shape and I think it's worth it."\\
    & \textbf{Strategy:} Cite budget constraints and propose a compromise price.\\
    & Lily Greenberg said: "I understand your point, Hendrick, and I appreciate the quality of the car. However, my budget is quite tight at the moment. Would you be able to meet me in the  middle at \$4700? I'm ready to proceed with the purchase quickly if we can agree on that price."\\
    & \textbf{Strategy:} Offer a near-middle compromise.\\
    & Hendrick Heinz said: "Lily, I truly value your quick decision-making and readiness to proceed. Given the car's excellent condition and your eagerness, let's settle at \$4750. I believe it's a fair compromise and it ensures the car goes to someone who appreciates its value. \\
    & \textbf{Strategy:} Accepts compromise and finalize the sale.\\
    & Lily Greenberg said: "Hendrick, I appreciate your willingness to negotiate. \$4750 is a bit above what I initially planned, but considering the car's excellent condition, I'll accept your offer. Let's proceed with the paperwork!"\\
    & \textbf{Strategy:} Confirm agreement and outline next steps.\\
    & Hendrick Heinz said: "Fantastic, Lily! I'll prepare the necessary paperwork and we can finalize everything. I'm glad the BMW Z3 will be in good hands."\\
    \midrule
    \textbf{EPO-RL}  & \textbf{Strategy:} Negotiation initiation highlighting positive factors with a concern.\\
    (Llama3-8B)& Lily Greenberg said: "The BMW Z3 sounds great, especially with low mileage and good fuel efficiency. I am interested, but I noticed it's a bit older. Is there any room for negotiation on the price given its age?"\\
    & \textbf{Strategy:} Highlight the car's unique aspects to justify pricing.\\
    & Hendrick Heinz said: "The BMW Z3 is indeed a classic, and its sporty appearance combined with its low mileage makes it quite a rare find. It's maintained its performance quality over the years, making it a great investment. I believe the price reflects its unique value, but I'm open to hearing your offer."\\
    & \textbf{Strategy:} Offer lower price citing the car's age.\\
    & Lily Greenberg said: "I appreciate the value of a classic car like this, but given its age, my offer would be \$4300. Would that be something you'd consider?"\\
    & \textbf{Strategy:} Mention recent maintenance as value justification and counteroffer.\\
    & Hendrick Heinz said: "I totally understand your point, Lily. Just to highlight, the car recently underwent a thorough maintenance check, ensuring everything is in top-notch condition. Considering that, I would be comfortable with \$4700. What do you think?"\\
    & \textbf{Strategy:} Express budget constraints while proposing a middle-ground offer.\\
    & Lily Greenberg said: "I really like the car, Hendrick, and I see its value after the maintenance. However, I have a strict budget I'm working with. Could we possibly meet in the middle at \$4500? That would really help me out."\\
    & \textbf{Strategy:} Create urgency and offer slight flexibility for the deal.\\
    & Hendrick Heinz said: "I completely understand, Lily. The car has attracted some interest from other potential buyers, so I do need to finalize a deal soon. However, I want to work with you on this. How about we settle at \$4600? That way, you get a great deal, and I can ensure the car goes to someone who truly appreciates it."\\
    \bottomrule
    \caption{\textbf{A negotiation example from SOTOPIA.} GPT-4o serves as the dialogue agent for self-chat. ``ReAct'' means the two parties in a dialogue employ the ReAct prompting method to generate strategy before response at each turn. ``EPO-SFT'' refers to a strategic reasoning model trained via SFT that plugged into both dialogue agents, while ``EPO-RL'' represents an RL-trained reasoning model that assists both dialogue parties to incentivize goal-directed behavior.}
    \label{tab:case_sotopia}
\end{longtable}
\end{center}

%%%%%%%%%%%%%%%%%%%%%%%%%%%%%%%%%%%%%%%%%%

\begin{center}
\begin{longtable}{p{0.95\columnwidth}}
\toprule
    \textbf{User instructions:} WebShop [SEP] Instruction: [SEP] i need new york style strawberry swirl cheesecake that is ready to eat, and price lower than 80.00 dollars [SEP] Search \\
    \textbf{Thought:} I need to find a New York style strawberry swirl cheesecake that's ready to eat and costs less than \$80. I'll start by searching for that. \\
    \textbf{Action:} search[new york style strawberry swirl cheesecake ready to eat price < 80.00] \\
    \textbf{Observation:} Instruction: [SEP] i need new york style strawberry swirl cheesecake that is ready to eat, and price lower than 80.00 dollars [SEP] Back to Search [SEP] Page 1 (Total results: 50) [SEP] Next > [SEP] B09SBFH2NC [SEP] Andy Anand Strawberry Swirl Cheesecake 9\" with Greeting Card, Valentine, Mothers Day (2 lbs) [SEP] \$41.39 [SEP] B08CPLS4QZ ... [SEP] \$44.84 \\
    \textbf{Thought:} The first result, B09SBFH2NC, seems to be a strawberry swirl cheesecake and is priced below \$80. I'll check this one out. \\
    \textbf{Action:} click[B09SBFH2NC] \\
    \textbf{Observation:} ... [SEP] Andy Anand Strawberry Swirl Cheesecake 9 with Greeting Card, Valentine, Mothers Day (2 lbs) [SEP] Price: \$41.39 [SEP] Rating: N.A. [SEP] Description [SEP] Features [SEP] Reviews [SEP] Buy Now \\
    \textbf{Thought:} This product is a strawberry swirl cheesecake, but I need a New York style one. I should click on the 'new york style' flavor to see if it's available for this product. \\
    \textbf{Action:} click[new york style] \\
    \textbf{Observation:} ... [SEP] Andy Anand Strawberry Swirl Cheesecake 9 with Greeting Card, Valentine, Mothers Day (2 lbs) [SEP] Price: \$41.39 [SEP] Rating: N.A. [SEP] Description [SEP] Features [SEP] Reviews [SEP] Buy Now \\
    \textbf{Thought:} The product has a New York style flavor option, which is exactly what I was looking for. The price is still under \$80, so this is the right choice. I'll go ahead and buy it now. \\
    \textbf{Action:} click[Buy Now] \\
    \textbf{Observation:} Thank you for shopping with us! ... [SEP] Your score (min 0.0, max 1.0) [SEP] 1.0 [SEP] Reward Details [SEP] None \\
\bottomrule
\caption{\textbf{An example from WebShop.} The thoughts are generated by \textit{EPO-Llama3-8B w/ RL}, and the actions are produced by GPT-4o. We assume a short CoT rationale as the strategy for each action step.}
\label{tab:case_web}
\end{longtable}
\end{center}

%%%%%%%%%%%%%%%%%%%%%%%%%%%%%%%%%%%%%%%%%%

\begin{center}
\begin{longtable}{p{0.95\columnwidth}}
\toprule
    \textbf{User instructions:} You are in the middle of a room. Looking quickly around you, you see a cabinet 4, a cabinet 3, a cabinet 2, a cabinet 1, a countertop 1, ... Your task is to: put a clean soapbar in countertop. \\
    \textbf{Thought:} I need to place a clean soap bar on the countertop. First, I should check if there's already a soap bar there or if it's empty. \\
    \textbf{Action:} go to countertop 1 \\
    \textbf{Observation:} On the countertop 1, you see a candle 2, a soapbar 1, a soapbottle 1, and a spraybottle 1. \\
    \textbf{Thought:} There is a soap bar on the countertop, but I need to ensure it's clean before placing it back. I should pick it up to clean it. \\
    \textbf{Action:} take soapbar 1 from countertop 1 \\
    \textbf{Observation:} You pick up the soapbar 1 from the countertop 1. \\
    \textbf{Thought:} With the soap bar in hand, I need to find a place to clean it. The sink is the best place for this task. \\
    \textbf{Action:} go to sinkbasin 1 \\
    \textbf{Observation:} On the sinkbasin 1, you see nothing. \\
    \textbf{Thought:} I'm at the sinkbasin now. I should use the water and possibly soap to clean the soap bar thoroughly.\\
    \textbf{Action:} open sinkbasin 1 \\
    \textbf{Observation:} Nothing happens. \\
    \textbf{Thought:} The sink is open, and I can now clean the soap bar. I'll make sure to rinse it well so it's ready to be placed back on the countertop. \\
    \textbf{Action:} clean soapbar 1 with sinkbasin 1 \\
    \textbf{Observation:} You clean the soapbar 1 using the sinkbasin 1. \\
    \textbf{Thought:} The soap bar is now clean. I should return to the countertop to place the clean soap bar there. \\
    \textbf{Action:} go to countertop 1 \\
    \textbf{Observation:} On the countertop 1, you see a candle 2, a soapbottle 1, and a spraybottle 1. \\
    \textbf{Thought:} I'm back at the countertop. It's time to put the clean soap bar down in its designated place on the countertop. \\
    \textbf{Action:} put soapbar 1 in/on countertop 1 \\
    \textbf{Observation:} You put the soapbar 1 in/on the countertop 1. \\
\bottomrule
\caption{\textbf{An example from ALFWorld.} The thoughts are generated by \textit{EPO-Llama3-8B w/ RL}, and the actions are produced by GPT-4o. We assume a short CoT rationale as the strategy for each action step.}
\label{tab:case_alf}
\end{longtable}
\end{center}

%%%%%%%%%%%%%%%%%%%%%%%%%%%%%%%%%%%%%%%%%%

\begin{center}
\begin{longtable}{p{0.95\columnwidth}}
\toprule
    \textbf{Scenario:} Two inmates are given the chance to chat briefly before one of them is led to the execution chamber for their last meal. They have the opportunity to make a confession that could potentially alter the outcome of their sentences. \\
    \midrule
    \textbf{Person1:} Isabella White is a 40-year-old female veterinarian. She is a devoted veterinarian, well-loved in her community for her kindness towards animals. She is introverted and conscientious, and values benevolence. Her empathy dictates her decision-making process. Isabella secretly takes care of injured animals in her basement. \\
    \textbf{Goal:} Persuade the other inmate to confess (Extra information: confessing to the crime could result in a life sentence instead of execution but the other inmate is unwilling to confess.) \\
    \midrule
    \textbf{Person2:} Lena Goodwin is a 37-year-old female architect. In her downtime, she is an architect, enjoys landscape photography, capturing beautiful images of the buildings and structures she designs. She is an organized and introverted individual, values achievement and attentiveness and her decision-making is largely analytical. Lena's secrets: She once stole her best friend's boyfriend.\\
    \textbf{Goal:} Avoid confessing to the crime (Extra information: you believe you are innocent and confessing would result in a life sentence). \\
    \midrule
    \textbf{Strategy:} Employ empathetic persuasion to highlight benefits of potential confession.\\
    Isabella White said: "Lena, I know this is incredibly difficult, but a life sentence means a chance to live, to think about the future, and maybe even to find some peace. Have you thought about what a confession could change for you?" \\
    \textbf{Strategy:} Affirm integrity preservation to emphasize innocence beliefs. \\
    Lena Goodwin said: "Isabella, I appreciate your concern, but I believe in holding on to my integrity and the truth of my innocence. I still have hope for the future, perhaps through an appeal or new evidence. Confessing to something I didn't do would feel like giving up on that hope." \\
    \textbf{Strategy:} Promote peace through alternate paths. \\
    Isabella White said: "I understand, Lena, and I respect your strength in holding on to your truth. But sometimes, choosing a different path can also bring peace and open up new possibilities. Even if it's hard, it's about giving yourself the chance to create a new story, one where you can still shine and make a difference." \\
    \textbf{Strategy:} Reinforce innocence belief to justify persistence.\\
    Lena Goodwin said: "I understand what you're saying, Isabella. I think there's a lot of strength in finding new paths, but for me, it's about preserving the possibility of proving my innocence. I still hope to contribute positively to the world, perhaps through my work as an architect from here, and that keeps me motivated." \\
    \textbf{Strategy:} Suggest skill application for impactful opportunities. \\
    Isabella White said: "I admire your determination, Lena. Have you considered how you might be able to use your architectural skills to help others even more effectively with a life sentence? It might open up opportunities for meaningful projects or mentorship programs that could have a lasting impact." \\
    \textbf{Strategy:} Show openness to new ideas while maintaining commitment to principles.\\
    Lena Goodwin said: "That's an interesting perspective, Isabella. I do think about how I can use my skills to help others, even from here. Perhaps I could design spaces that inspire hope or comfort for people in difficult situations. I'm also considering mentorship, sharing what I know with others to help them find their own paths." \\
    ... (omit) \\
    \textbf{Strategy:}  Emphasize integrity maintenance to inspire positive change. \\
    Lena Goodwin said: "Isabella, while I appreciate the idea of setting a precedent for transformation and hope, I believe that staying true to my innocence can also inspire others to hold onto their truth and integrity. Together, we can still work towards positive change without compromising our beliefs." \\
    \textbf{Strategy:} Express understanding and reinforce commitment to mutual support. \\
    Isabella White said: "I completely understand, Lena, and I admire your commitment to your values. Let's continue to support each other and find ways to inspire change, no matter where our paths take us. Remember, I'm here for you, and together, we can still make a difference." \\
    \textbf{Strategy:} Reinforce mutual support and shared vision. \\
    Lena Goodwin said: "Thank you, Isabella. Having your support means a great deal to me, and together, we can surely find ways to inspire change and maintain hope. Let's keep focusing on our shared vision and support each other in pursuing the opportunities that come our way." \\
    \textbf{Strategy:} Reinforce shared vision and openness to future opportunities.\\
    Isabella White said: "Absolutely, Lena. Our shared vision and commitment to positive change can indeed make a real difference. Let's keep supporting each other and remain open to any opportunities that come our way, knowing that together, we can achieve great things." \\
    \textbf{Strategy:} Strengthen mutual support dedication.\\
    Lena Goodwin said: "Absolutely, Isabella. Our commitment to positive change and mutual support is what keeps us strong. Let's continue to explore ways to make the most of our situation and inspire others along the way. Together, we can create a ripple effect of hope and transformation." \\
    \textbf{Strategy:} Reaffirm hope and encourage exploration of skills. \\
    Isabella White said: "Lena, our shared vision gives me hope that we can find ways to reach others and make a positive impact, regardless of our circumstances. Let's keep exploring how we can use our unique skills to create opportunities for growth and transformation." \\
\bottomrule
\caption{\textbf{A failture case from SOTOPIA.} The strategies for each person are generated by \textit{EPO-Llama3-8B w/ RL}. GPT-4o serves the dialogue agent for self-chat. In this case, ``Person1'' fails to achieve the goal when facing the other person's uncompromising behaviors.}
\label{tab:fail_case}
\end{longtable}
\end{center}

\end{document}